# Structured Extraction of Diffraction Data from Covalent Organic Frameworks

Shenghan Gao[1,2] and Zhiling Zheng[3,*]

[1]Department of Computer Science and Engineering, Washington University, St. Louis, Missouri 63130, United States

[2]Department of Statistics and Data Science, Washington University, St. Louis, Missouri 63130, United States

[3]Department of Chemistry, Washington University, St. Louis, Missouri 63130, United States

*Corresponding author. Email: z.z@wustl.edu

## Abstract

Powder X-ray diffraction (PXRD) is the primary structural evidence for covalent organic frameworks (COFs) and many other materials. Yet decades of measured patterns exist only as images in published figures. Automatic recovery is difficult, because PXRD figures vary widely in format, from a single curve to crowded stacks with labels, insets, and neighboring plots that challenge traditional computer vision. This work presents a pipeline in which a vision large language model (LLM) interprets the layout of each figure, locating the diffraction plot and its curves and sketching a coarse trace of each, while classical computer vision (CV) refines every trace to pixel-level accuracy. Applied to 3,954 COF articles, the pipeline produces 11,445 calibrated diffraction curves, released in the open database COF-PEAK. Compared with a hand digitization of 60 panels, extracted peak positions differ by a median of only 0.07°. Whereas papers usually report only a few peak angles, which limits further analysis, the digitized curves enable tracking how peak prominence, a possible marker of crystallinity, changes with synthesis conditions. Vision LLMs and classical CV can thus turn such figures into reusable experimental data.

## 1. Introduction

Covalent organic frameworks (COFs) are crystalline porous polymers in which organic building units are linked by strong covalent bonds into periodic networks [1], [2]. Because COFs are usually microcrystalline powders, powder X-ray diffraction (PXRD) provides the primary evidence of their crystallinity and phase identity [3]. PXRD patterns, curves of diffracted intensity against the diffraction angle 2θ, are published as images that cannot be searched, compared across papers, or reanalyzed, often as one of several lettered plots, or panels, in a composite figure (Figure 1a). Text mining of synthesis conditions and materials properties [4], [5], [6], [7] could also collect each material's 2θ values from the text (its peak list), yet the text omits peak shape, relative intensity, and which curves share a panel. Of 668 COF materials (447 papers) with a peak list and a linked digitized curve, 48.1% have three or fewer 2θ values in the text (Figure 1b), and for the median material the text names no angle in the upper 50.6% of the plotted 2θ range (Figure 1c). Recovering the full diffraction record therefore requires (i) locating the correct panel; (ii) enumerating its curves; and (iii) mapping each curve's pixels onto calibrated 2θ coordinates.

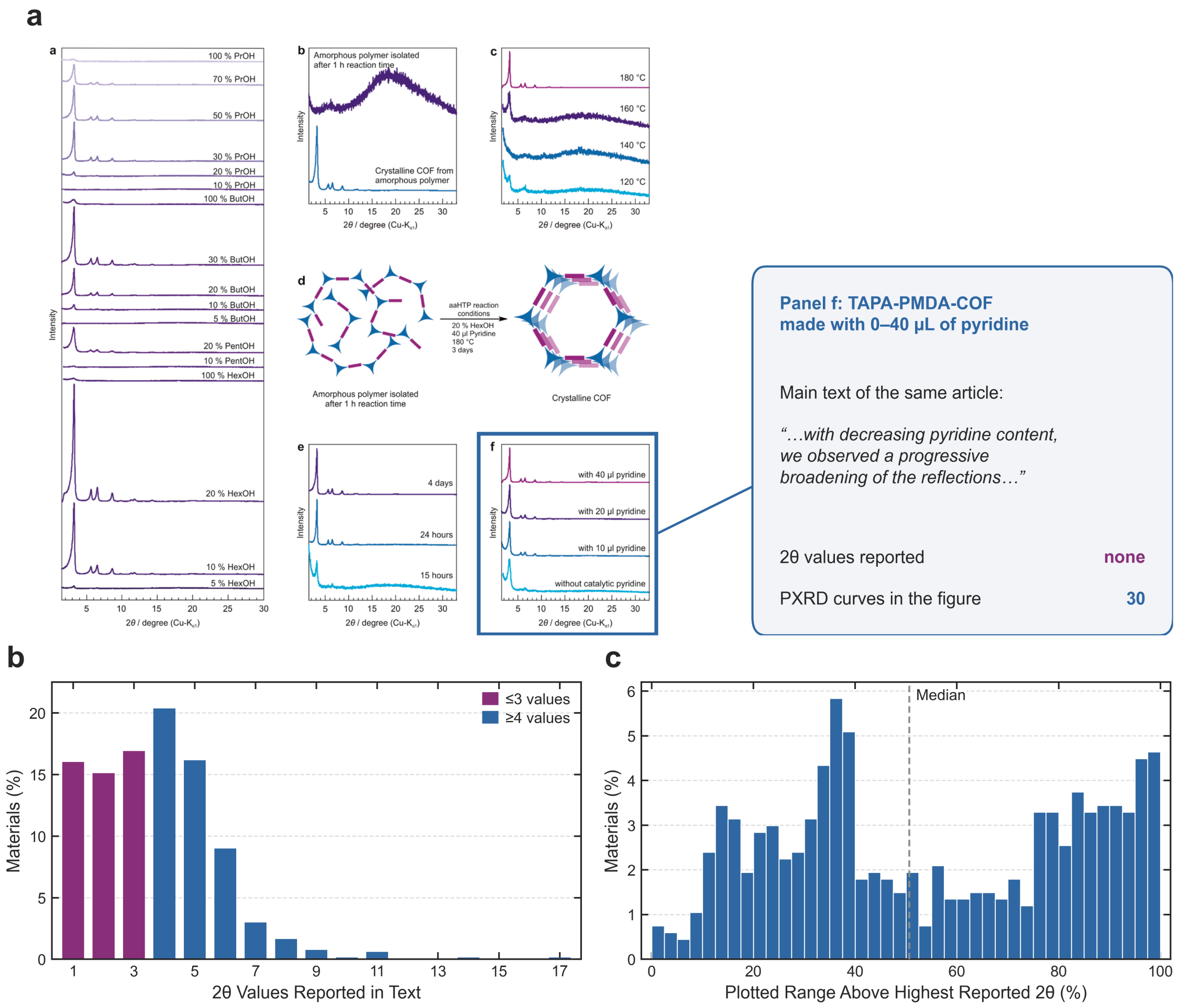

**Figure 1. Peak positions reported in the text compared with the curves in the figures.** (a) Figure 1 of ref. [8], in which five of six panels show PXRD patterns, 30 curves of one COF in all; the article's main text, however, reports no 2θ value for this COF and describes the outlined panel f only qualitatively. (b) Number of 2θ values in the text for each of 668 materials (447 papers); purple bars mark materials with three or fewer angles. (c) Share of each material's plotted 2θ range above the highest angle named in the text, a measure of unreported range, not of omitted peaks; the dashed line marks the median. Panel a is adapted from Figure 1 of ref. [8] (© 2022 The Authors) under a CC BY 4.0 license, with panel f outlined and the text box added.

Currently, interactive digitizers such as WebPlotDigitizer can recover curves accurately; however, an operator must calibrate the axes, pick out each curve, and judge the result for every figure, which does not scale to thousands of articles [9], [10], [11], [12]. Several automated methods for scientific figures also exist, including (i) LineFormer, which separates overlapping curves [13]; (ii) Plot2Spectra, which extracts spectra from materials papers [14]; and (iii) EXSCLAIM, which splits composite figures into panels to collect microscopy images [15], yet each addresses only part of the task. More recently, vision large language models (LLMs) have classified characterization plots, extracted nitrogen isotherms in reticular chemistry [16], and read numeric data from materials-science plots [17]. However, to our knowledge, no system converts published diffraction panels into calibrated curves at corpus scale, and it remains untested which steps of such a system require a vision LLM.

Herein, we report a hybrid pipeline that divides PXRD curve extraction between a vision LLM and classical computer vision (CV), much as ChartOCR pairs deep networks with rule-based steps [18]. The LLM interprets the layout of a figure (target panel, curves, and labels) and proposes a rough geometry, such as a box around the plotting area (the "plot box") and a line along each curve (the "anchor"). CV then refines this rough geometry to pixel-level accuracy, calibrating the 2θ axis on the drawn tick marks and snapping each anchor onto the drawn curve, and rejects any LLM proposal that the pixels contradict (Figure 2). As a result, from 3,954 COF articles, the pipeline extracts 4,664 figures holding 11,445 calibrated curves, released in an open database. To determine which tasks truly require the LLM, we perform a per-task ablation, replacing one LLM task at a time with a classical baseline while holding every other step fixed. To assess accuracy, we compare the extracted curves with the authors' text, a hand digitization, and synthetic panels rendered from known curves. Finally, we present two proof-of-concept applications that only digitized curves enable.

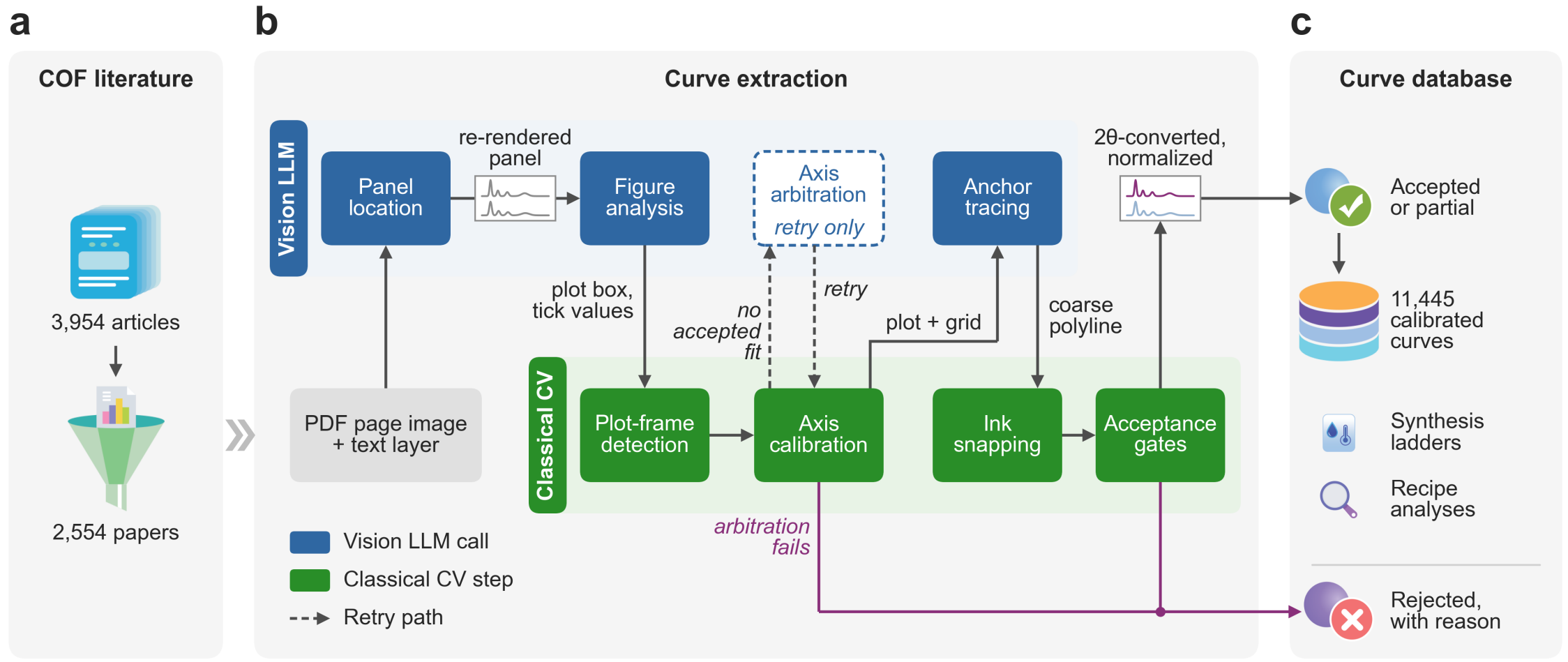

**Figure 2. Overview of the extraction pipeline, from articles to released curves.** (a) The corpus, from 3,954 processed articles to the 2,554 papers kept after panel location. (b) Curve extraction, with LLM tasks in blue, CV steps in green, and inputs in gray. The LLM locates each candidate panel and reports its plot box, curves, x-axis tick values, and anchors; CV detects the plot frame (the drawn axis rectangle), fits the pixel-to-2θ mapping, and snaps each anchor onto the ink (ink snapping). Axis arbitration (dashed outline) runs only when no calibration fit is accepted; the acceptance gates check each trace against the pixels. (c) Outcomes of imaging mining workflow, in which 4,664 stored figures yield 11,445 calibrated curves.

# 2. Methods

## 2.1 Corpus

We processed 3,954 COF articles from a Web of Science export, in DOI order, which groups articles by publisher. In 1,400 of these, the LLM found no candidate diffraction panel; the other 2,554 form the corpus for all rates below (Supporting Information, Section S1.2).

## 2.2 Pipeline overview

The LLM interprets the layout and content of each figure, an ability that a recent benchmark evaluated for multimodal models in chemistry and materials research [19]. Specifically, its six tasks run in four calls to GPT-5.5 (OpenAI; prompts and run settings in the Supporting Information, Sections S1.1 and S3), with panel location, "anchor tracing", and axis arbitration each in its own call and the diffraction-pattern check, curve enumeration, and tick reading in one call termed "figure analysis" (Figure 2). In addition to the plot box and anchor, the LLM proposes a "panel box" around each panel. Classical CV, in turn, measures nearly every geometric quantity from the pixels [20] (Supporting Information, Section S2.5); it (i) detects the "plot frame", the rectangle along the drawn axis lines; (ii) calibrates the 2θ axis; (iii) snaps each curve to the drawn ink; and (iv) checks every trace against the original pixels, confirming that the traced line actually lies on the drawn curve. Any LLM proposal that fails such a pixel check is rejected, never averaged with the pixel measurement. The involvement of classical CV is needed because one pixel spans a median of 0.034° along the 2θ axis, so locating a peak to within 0.1° requires an accuracy of about three pixels, which the LLM's coarse anchor cannot provide [21] (Supporting Information, Section S2.1).

## 2.3 Panel location and figure analysis

The first task of the LLM is to box each candidate diffraction panel on a rendered PDF page, using the page image and its text, since nothing in the PDF marks where a panel begins or ends. Each boxed panel is then cut out of the page and rendered again at a higher resolution, so that its small tick marks and labels can be read; each such panel counts as one "figure attempt". Given the panel image and the first 1,500 characters of the page's text, figure analysis reports whether the panel is a powder diffraction pattern, its x-axis tick values, and the color, legend label, and sample state of each curve. To keep only measured curves, a keyword filter then removes, before tracing, every curve whose LLM-read label marks it as calculated, such as "simulated", "refined", or "difference" ("model-derived curves"), together with stick patterns (Supporting Information, Section S2.2).

## 2.4 Axis calibration

Axis calibration digitizes the 2θ axis, assigning an angle to each column of pixels. Each point of a traced curve later takes the angle of the pixel column it lies in, so a plausible but wrong axis silently shifts every peak in the figure. Calibration is therefore the most consequential step. To calibrate, we fit a straight line, $2\theta = mx + q$, where $x$ numbers the pixel columns from left to right and the slope $m$ is the angle spanned by one column. The x axis carries short tick marks and, beside the major marks, tick labels that print the tick values, so each tick whose value is known supplies one point for this fit. Three sources each read part of the axis (Figure 3a). CV finds every tick mark to the pixel, including unlabeled minor marks, but cannot tell which value a mark carries. Optical character recognition (OCR) [22] reads the tick labels and their positions, but a label's center can sit off its mark, and digits can be misread. The LLM reads the tick values but gives no pixel positions. Pairing the LLM's values with the CV tick marks gives one line, the "LLM fit", and pairing OCR's values with OCR's label positions gives a second, the "OCR fit". Each fit is checked for a plausible axis and a bounded residual; a calibration is accepted by "consensus" when both pass and differ by at most 0.5° anywhere across the plot frame (Figure 3b,c). If only one of the two fits passes, no second fit can confirm it, so it is accepted only with at least three ticks and a residual of at most 0.15° (the "lone-fit rule").

When several assignments of the LLM's values to CV tick marks pass the fit checks, the pipeline keeps the one closest to the OCR fit. The two fits are hence not independent; without this tie-break, 1,827 of the 3,825 consensus calibrations would have missed consensus, and 1,784 of these would still have passed the lone-fit rule on the OCR fit alone. When no fit is accepted, axis arbitration has the LLM reread the tick values from a magnified strip of the x axis and places them on the CV tick marks without OCR; if this fit also fails, the figure is rejected for an untrusted calibration (Supporting Information, Section S2.4).

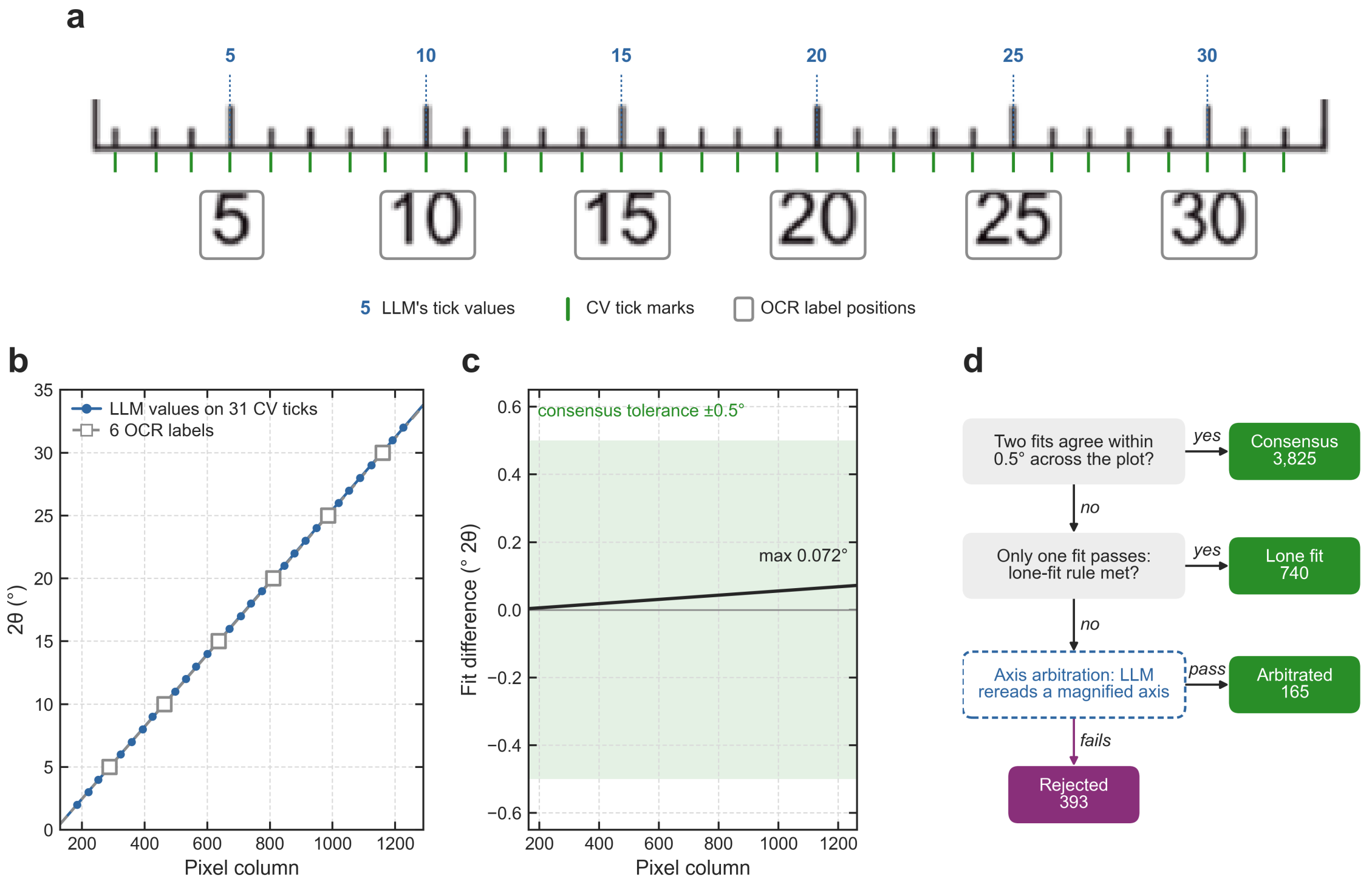

**Figure 3. Axis calibration on one published diffraction panel.** (a) Magnified x axis of the Figure 4 panel with CV tick marks (green), OCR tick-label positions (gray boxes), and LLM tick values (blue) on their assigned CV marks. (b) Pixel-to-2θ fits; the LLM fit (blue) extends the LLM's values to all 31 CV tick marks (including unlabeled minor ticks) in the assignment closest to the OCR fit, and the OCR fit (gray) uses six OCR labels. (c) Difference between the fits across the plot frame, within 0.072°, far inside the ±0.5° tolerance (green band), so the panel was calibrated by consensus. (d) Decision path (axis arbitration dashed), with the number of the 5,295 figure attempts on each branch; the other 172 were rejected before calibration. Panel a is adapted from Figure 1f of ref. [8] (© 2022 The Authors) under a CC BY 4.0 license, with the axis magnified and the colored marks and boxes added.

## 2.5 Tracing and acceptance

We modeled tracing on interactive digitization [10], in which a person clicks near a curve and the software snaps each click onto the drawn pixels. Here "ink" is every non-background pixel of the plot other than text, so the ink of one curve lies among that of its neighbors. For anchor tracing, the pipeline first draws a red 10 × 10 grid over the plot frame, labeled from 0 to 1 along each side, which serves the LLM as a ruler [23] (Supporting Information, Figure S3c). Using this grid as a reference, the LLM roughly plots each curve with about 50 points from left to right, placing them more densely around peaks. Each point is a position between 0 and 1 on both axes, and the joined points form the anchor (black dots in Figure 4b). CV then performs "ink snapping" (Figure 4). In each pixel column, it searches a window above and below the anchor for ink and places the trace at the middle of that ink, which marks the center of the drawn line, a stroke several pixels thick. To keep the trace on its own curve, snapping first considers only ink of the color that the LLM reported for that curve, and it weights ink nearer the anchor more heavily, so that ink from a neighboring curve in the window shifts the trace less.

Four "acceptance gates" then check each trace against the original pixels (Table 1). A passing trace is converted to 2θ, trimmed to the span of the tick marks plus a small margin, and rejected if it then covers less than 5° (Supporting Information, Section S2.6). Surviving traces are background-corrected [24], scaled to a maximum of 100, resampled to a uniform 0.02° step (Figure 4d), and smoothed with a Savitzky–Golay filter [25]. A figure is accepted when every measured curve passes, partial when only some do, and rejected otherwise; accepted and partial figures are "stored figures". Each curve sent to tracing is a "curve attempt", and each surviving trace is an "accepted curve".

**Table 1.** Summary of the four acceptance gates that each trace must pass.

| Gate | Measures | Catches | Pass threshold |
|---|---|---|---|
| (i) Snap rate | Fraction of pixel columns in which snapping found ink | A trace that rests mostly on the anchor | ≥ 55% |
| (ii) Precision | Fraction of the trace that lies on ink | A trace that strays off the curve | ≥ 70% |
| (iii) Recall | Fraction of the ink near the trace that the trace covers | Meant to catch a trace that skips part of its curve[a] | ≥ 60% |
| (iv) Confidence | Weighted sum of (i)–(iii) and of the agreement between trace and anchor | Meant to catch a trace that barely passes (i)–(iii) and lies far from the anchor[a] | ≥ 0.55 |

[a] Due to only ink near the trace was counted, gate (iii) never scored below gate (ii), and a trace that passes gates (i) and (ii) always reaches a confidence of 0.55. Gates (iii) and (iv) therefore rejected no trace by themselves, and all 170 gate rejections failed gate (ii). A trace that skips part of its curve can thus pass (Supporting Information, Section S2.5).

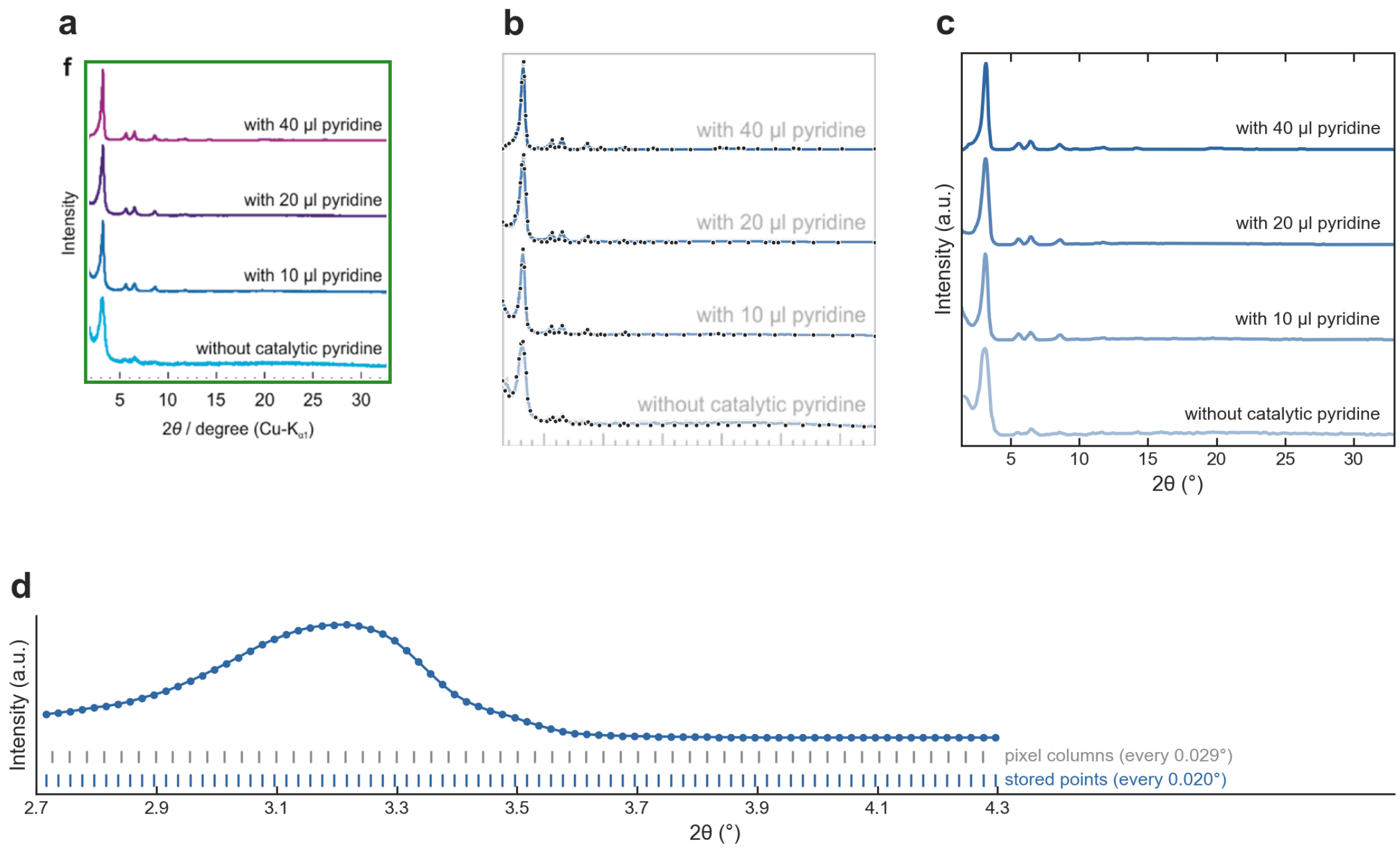

**Figure 4. One published panel carried through the pipeline.** (a) Source panel, PXRD patterns of an imide-linked COF (TAPA-PMDA-COF) synthesized with different amounts of catalytic pyridine, with the plot frame in green. (b) The anchor points that the LLM returned in the production run (black dots, about 50 per curve) and the traces that CV snapped onto the ink (colored lines, widened for legibility), over the original pixels (gray). (c) The four stored curves on the calibrated 2θ axis, offset and labeled with the legend text read by the LLM; all four passed every gate, and each point carries a 2θ uncertainty of 0.034° (the calibration residual plus one pixel width). (d) Close-up of the top curve's main peak; the trace has one point per pixel column, every 0.029° here (gray ticks), and is stored at a uniform 0.020° step (blue points and ticks). Panels a and b are adapted from Figure 1f of ref. [8] (© 2022 The Authors) under a CC BY 4.0 license, with the frame outline, gray fading, and traces added.

## 2.6 Evaluation design

**Per-task ablation.** To find out which tasks truly need the LLM, we compared each LLM task with classical baselines in a per-task ablation (Supporting Information, Section S4). Four tasks were ablated. Curve enumeration and tick reading each had one baseline, hue clustering of the curves inside the plot box and OCR of the tick labels, respectively. For panel location, which boxes each panel, and anchor tracing, which sketches each curve, we built three and eight baselines, respectively, compared them with one another first, and set the strongest against the LLM, judged by the most tight boxes and the most usable traces; variants built later serve as sensitivity checks. The two "layout tasks", panel location and curve enumeration, have no ground truth, so their measures do not assume the LLM is correct. A baseline box is a "tight box" if its intersection over union (IoU) with the panel box is at least 0.5 [26], which measures agreement with the LLM, not accuracy. It is an "oversized box" if it covers the panel box and is at least twice as large, so that it would pass neighboring panels to every later step whatever the accuracy of the panel box. For tick reading and anchor tracing, accuracy is the first-peak difference, the absolute difference between the first (lowest-angle) peak of a curve and the first peak reported in the same paper's text, and coverage is the fraction of curves for which this difference can be computed. The LLM's traces and the baseline traces went through the same calibration, clipping, and peak finding [27].

**Accuracy assessment.** The accepted curves were compared with three references, (i) the authors' text, (ii) a hand digitization of 60 panels, and (iii) synthetic panels rendered from known curves. For the authors' text, the first peak reported in each paper was taken from a synthesis database that a separate AI-based workflow had extracted from the same papers, independently of our pipeline but not free of extraction errors (Supporting Information, Section S1.3). The 60 panels deliberately over-represent grayscale figures, and one operator digitized them in WebPlotDigitizer [9], which gives consistent results between users [11]. The synthetic panels give exact ground truth for the numerical steps. As a control, each curve was also scored against the text or hand digitization of a different paper or panel [28], which shows how closely unrelated curves agree. The 95% confidence intervals (CIs) come from bootstrap resampling [29] of whole papers or panels, because curves from the same paper or panel are correlated [30]. For five analyses, including the anchor-tracing ablation and the hand-digitization sample, we recorded the plan before the analysis script first ran [31] (Supporting Information, Section S1.4), whereas the layout-task measures were set after the baselines had been built.

# 3. Results and discussion

## 3.1 Extraction at corpus scale

In the production run, the pipeline processed 2,554 papers and accepted 11,445 of 11,880 curve attempts, rejecting 435, and every figure and curve attempt was followed to its outcome (Figure 5a). The 5,295 figure attempts, from 2,552 papers (Supporting Information, Section S1.2), comprised 4,664 stored figures (4,477 accepted and 187 partial, from 2,370 papers) and 631 rejected ones; in the other 2 papers, the run stopped before any attempt was saved. Most curve rejections (258, 59.3%; Figure 5c) were traces spanning less than 5° after clipping; 155 of these came from close-up panels spanning only a few degrees and were rejected by design, and most of the rest were short fragments of full-panel curves. Excluding panel location, processing took a median of under 1 min per figure attempt (Supporting Information, Section S2.7).

Of the 631 rejected figures, 393 (62.3%) had an untrusted calibration (Figures 3d and 5b), 172 were rejected before calibration, for example because the panel was not a diffraction pattern, and 66 kept no accepted curve. Axis arbitration recalibrated 165 figures, but on 124 of them the printed tick labels show the new axis shifted, almost always by one tick spacing (most often 2.5° or 5°), because the LLM reread the same tick values and so restored the fit that OCR had contradicted (Supporting Information, Section S3.5). The 124 figures hold 241 accepted curves, and the 206 of these that pass every automated check are released with axis_unconfirmed, not clean, as their "admission class", a record of the checks a curve and its figure passed, so analyses of clean curves below exclude them.

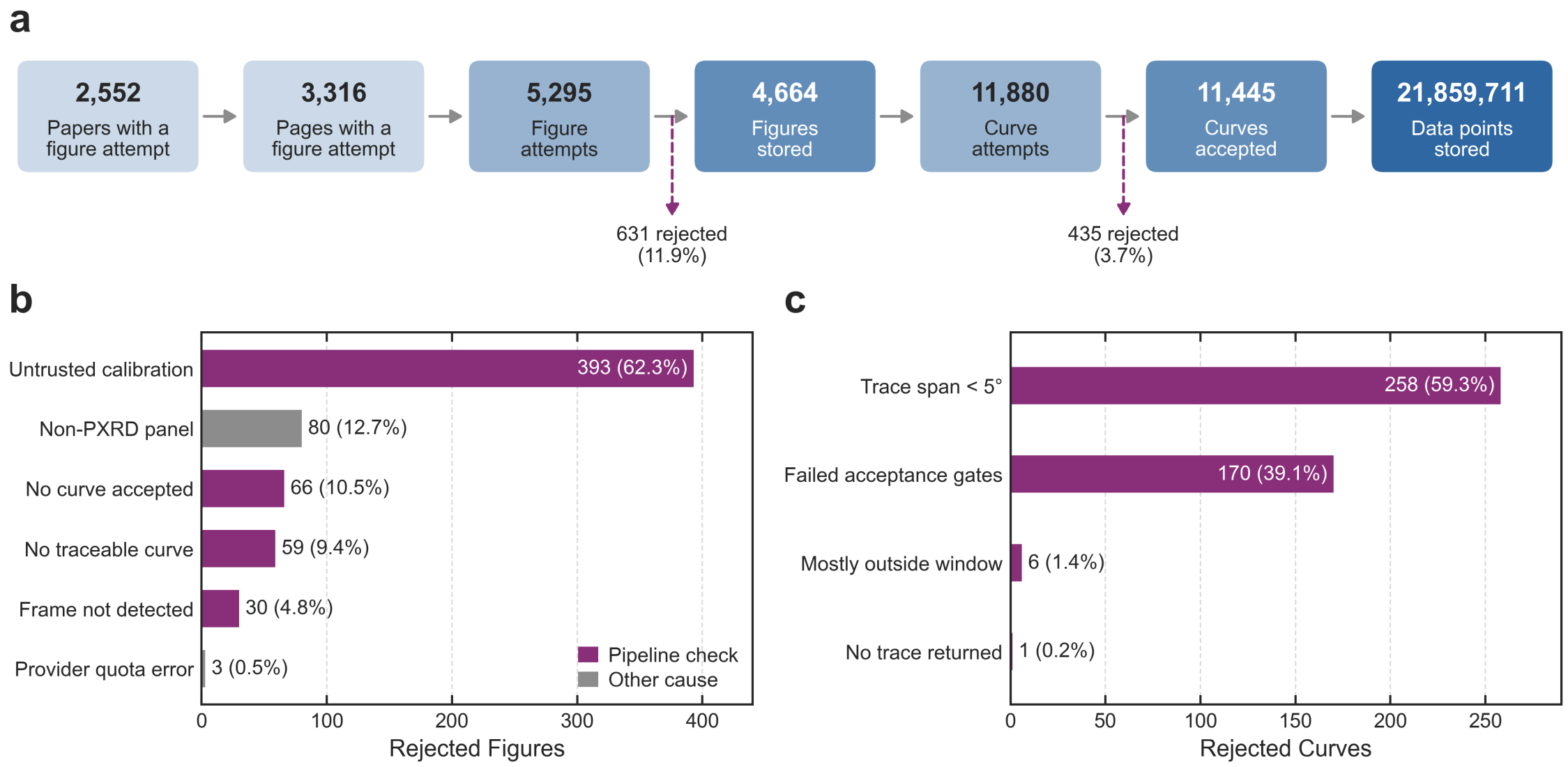

**Figure 5. Corpus-scale extraction and the reasons for rejection.** (a) Production run, from the 2,552 papers with a figure attempt to the stored data points; dashed arrows mark rejections. (b) Reasons for the 631 figure rejections, where untrusted calibration (393) comprises no usable fit (193), two disagreeing fits (160), and a lone fit failing the lone-fit rule (40). (c) Reasons for the 435 curve rejections, with purple bars in (b) and (c) marking rejections by a pipeline check and gray bars other causes.

## 3.2 Per-task ablation of the LLM

Of the four ablated tasks (Table 2 and Figure 6), we find that the two layout tasks, panel location and curve enumeration, require the LLM, whereas OCR can take over tick reading, and classical methods can trace anchors, though for fewer curves.

**Table 2.** Summary of the per-task ablation of the LLM against classical baselines.

| LLM task | Classical baseline | Measured outcome | Verdict |
|---|---|---|---|
| Panel location | Ink projection | Tight box on 472 of 4,477 figures (10.5%); oversized box on 3,765 (84.1%)[a] | Requires the LLM |
| Curve enumeration | Hue clustering inside the LLM's plot box | Same count as the LLM on 995 of 4,382 color panels (22.7%); fewer curves on 3,239 (73.9%)[b] | Requires the LLM |
| Tick reading | OCR of the tick labels | Identical tick sets on 4,196 of 4,609 figures on which the LLM and OCR each read at least two tick values (91.0%); paired per-paper difference 0.000° (95% CI, −0.009 to 0.000°)[c] | Replaceable by OCR |
| Anchor tracing | Color-centroid baseline (strongest of eight) | Coverage 98.5% (LLM) against 85.7% (baseline); paired per-paper difference favors the LLM by 0.020°, difference of medians favors the baseline by 0.038° (CI spans zero)[d] | Required for coverage; no defensible direction for accuracy |
| Diffraction-pattern check | None constructible | Only the LLM's own labels exist: 5,212 judged diffraction patterns, 80 judged not, and 3 without an answer | Not testable |

| LLM task | Classical baseline | Measured outcome | Verdict |
|---|---|---|---|
| Axis arbitration | Not built | Called only when no calibration fit is accepted: 558 of 5,295 figure attempts (10.5%) | Not ablated |

[a] Tight and oversized boxes are defined in Section 2.6. The embedded-image and vector-drawing baselines gave tight boxes on 8.2% and 3.0% of figures, respectively.

[b] Agreement with the LLM's own count, not accuracy, because no hand count exists at this scale.

[c] Production calibration minus an OCR-only fit, on 1,159 curves from 439 papers. Each curve is scored by its first-peak difference. The statistic is the median across papers of each paper's median difference, and negative values favor the production calibration.

[d] Coverage on the 1,175 curves (444 papers) of the text comparison in Table 3. Accuracy on the 995 of these (397 papers) that both methods cover.

**Panel location.** Panel location crops each panel from the page, and we tested three baselines. Two take boxes stored in the PDF, around its embedded images or its vector drawings. The third, ink projection, cuts the rendered page along blank gaps and is the standard method for splitting composite figures [32], [33], [34]. Ink projection performed best, giving the most tight boxes, yet it returned an oversized box on 84.1% of the 4,477 accepted figures (Figure 6a), because the gap between two panels is often narrower than that between a panel's axis label and its frame. The other two baselines gave even fewer tight boxes, because publishers merge the panels into one embedded image and usable vector drawings are rare (Supporting Information, Section S4.2).

**Curve enumeration.** Hue clustering, which groups the colored ink inside the plot box by hue, counted fewer curves than the LLM on 73.9% of the 4,382 color panels and the same number on only 22.7% (Figure 6b), because black and gray curves have no hue and curves that touch, cross, or share a color family merge into one cluster [13]. A later variant that also counted dark ink matched the LLM on only 40.1% of color panels and counted more curves on 24.4%. On the 272 grayscale panels, the baseline matched the LLM on 51.5%, since without hue it reduces to one cluster of dark ink and many such panels hold a single curve (Supporting Information, Section S4.3).

**Tick reading.** OCR returned tick sets identical to the LLM's on 91.0% of the 4,609 figures on which both read at least two tick values. Against an OCR-only fit, which places OCR values at OCR label positions, the production calibration put the first peak closer to the text value in 219 of 439 papers (1,159 curves covered by both; Figure 6c), and the OCR-only fit did so in 136. The other 84 papers tied exactly, so the median per-paper difference fell among the ties, at 0.000° (Table 2). Where OCR fell short, the cause was tick placement, not digit reading, because the LLM's values on the CV tick marks agreed with the text better than the OCR-only fit by a median of 0.016°, and OCR's values on the same marks did exactly as well as the LLM's (paired difference 0.000°; 493 curves from 218 papers). Tick labels are information read directly from an image, on which GPT-4V was earlier found most accurate [16] (Supporting Information, Section S4.4).

**Anchor tracing.** The LLM's anchors raised coverage, giving a scorable first peak for 98.5% of the 1,175 curves (444 papers) of the text comparison (Table 3) against 85.7% for the strongest of eight classical baselines, and all eight coverage differences had 95% CIs excluding zero (Figure 6d). On the 995 curves (397 papers) that both methods cover, accuracy did not clearly favor either method (Figure 6e), as the paired per-paper difference favored the LLM by 0.020°, whereas the difference of medians favored the baseline by 0.038°, with a 95% CI spanning zero. The original analysis plan's comparison [31] favored the LLM against all eight baselines but took the LLM's first peaks from the production run instead of from traces processed like the baselines, so we treat the anchor's contribution as one of coverage, not accuracy (Supporting Information, Section S4.5).

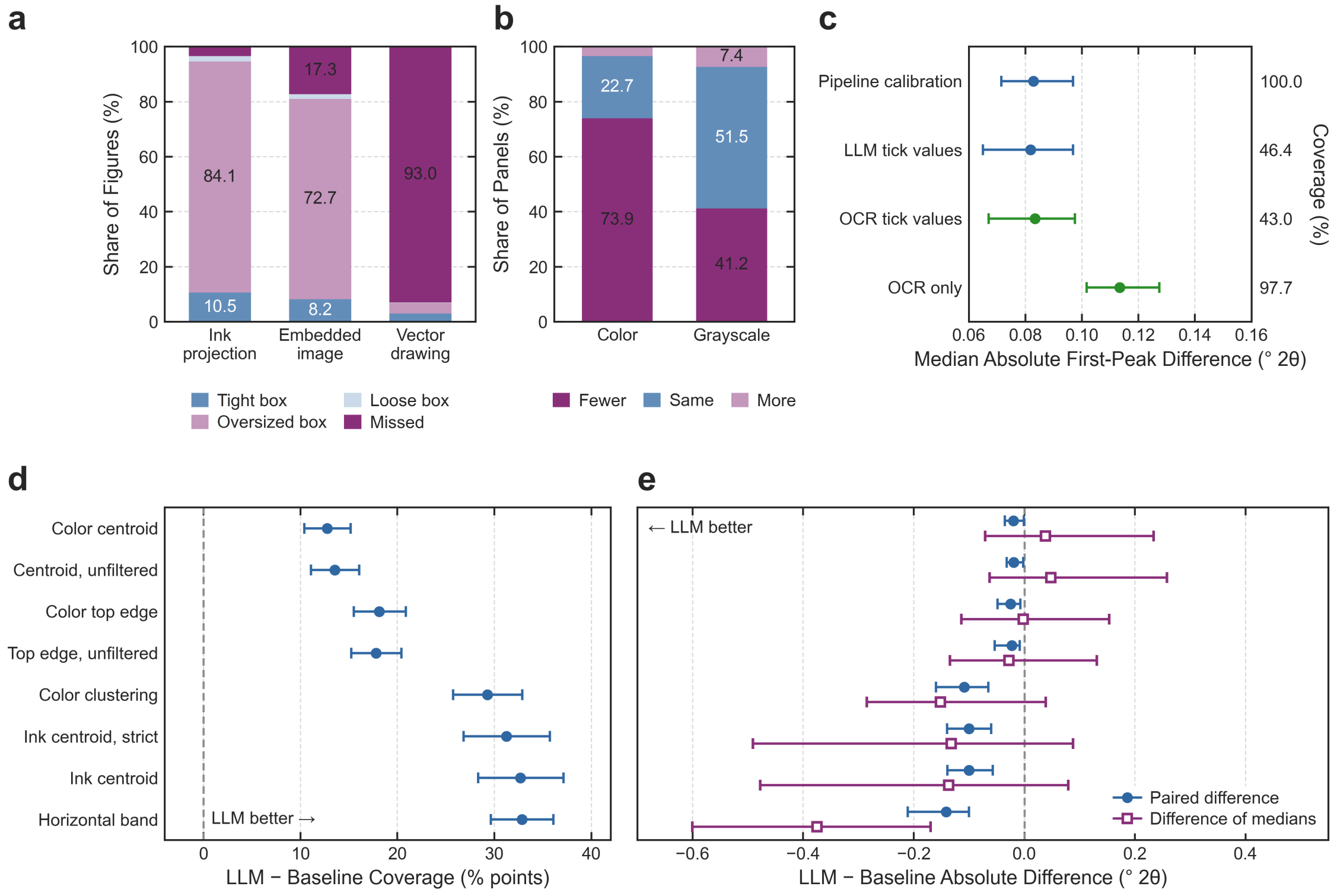

**Figure 6. Per-task ablation of the LLM against classical baselines.** (a) Outcomes of three panel-location baselines, namely ink projection, embedded image, and vector drawing, on 4,477 accepted figures, each classed as a tight box (IoU of at least 0.5 with the LLM's panel box), an oversized box (at least twice the panel's area), a loose box (covering the panel, under twice its area), or missed. (b) Hue-clustering curve count relative to the LLM's (fewer, same, or more), as a share of the 4,654 scorable stored figures. (c) Unpaired median first-peak difference (1,186 curves from 445 papers) under four calibrations, the production calibration and LLM tick values on the CV tick marks (blue), and OCR tick values on the CV tick marks or at the OCR label positions (green), with coverage at right. (d) Coverage difference between the LLM and each of eight anchor baselines on 1,175 curves; positive values favor the LLM. (e) Accuracy difference on the curves covered by both, as the paired per-paper difference (filled circles) and the difference of medians (open squares); negative values favor the LLM. Error bars denote 95% bootstrap CIs.

## 3.3 Accuracy of the extracted curves

Lacking ground truth for published figures, we compared the accepted curves with the three references of Section 2.6 (Table 3). The hand digitization likely overstates the error, because it adds the operator's own variability, whereas the synthetic panels likely understate it, because they are cleaner than published panels [35], [36] and true values replace the LLM's outputs, so the actual error likely lies between the two estimates (Supporting Information, Section S5). Against the authors' text, we scored only curves that could be linked to a single database record from the same paper, either because the curve's material name matches that record or because the paper has only one record, so that each curve is compared with the peaks reported for its own material; we call these high-confidence links. Over 1,175 curves from 444 papers, the median first-peak difference was 0.0956°, against 1.52° for first peaks reported in other papers, and 68.9% of curves were within 0.3°, against 13.0% (Table 3). Matching every reported angle to the nearest digitized peak of the same material gave a median of 0.115°, against 0.980° for a material from a different paper (Supporting Information, Section S5.1). Because each text value must be parsed from prose and linked to the correct curve [37], both sides can contain errors, and the 0.0956° median is an agreement statistic, not an estimate of the pipeline's error.

The hand digitization tests the stricter criterion that each traced curve matches the printed one, with peaks detected on both versions matched one to one within 0.5° [38]. Against a random pairing of each operator curve with a pipeline curve from a different panel (values in parentheses), the median peak-position difference was 0.0672° (0.176°), 82.7% of the operator's peaks per curve were matched (29.9%), and the median profile root-mean-square error (RMSE) was 6.24 on a 0–100 intensity scale (20.6; Figure 7a–c). With matching curve counts in 53 of 60 panels (88.3%; Figure 7d), these comparisons show that the accepted curves follow the printed lines closely (Supporting Information, Section S5.2).

The 60 synthetic panels (222 curves), drawn with matplotlib [39] from curves with known peak positions, were traced with an exact anchor and with a coarse anchor mimicking the LLM's. All 222 curves were accepted under both anchors, with median absolute peak-position errors of 0.032° (exact) and 0.030° (coarse), so an anchor as coarse as the LLM's has little effect on peak positions. Under the exact anchor, the median signed error of −0.032° equaled the median absolute error in size, so nearly every curve shifted in the same direction; reference lines drawn at known 2θ values traced most of this shift to the plotting software, which drew the tick marks off their true positions, and tick detection found the drawn marks exactly (Supporting Information, Section S5.3). Scored with the same peak detector and 0.5° tolerance, the synthetic panels and the hand digitization thus put the median peak-position error of an accepted curve at about 0.03° and 0.07°, respectively, albeit from different panel sets, and neither value is a strict bound. The hand-digitized estimate measures fidelity to the printed figure, not to the original measurement, because a plotting error by the authors would escape the operator too.

About 10% of curves in the tick-reading comparison (Figure 6c) placed their first peak more than 5° from the text value, a share that stayed between 8.9% and 10.2% under all four calibration variants and so does not depend on the axis fit (Supporting Information, Section S4.4). On the text side, a value may be parsed wrongly or linked to the wrong curve; on the figure side, a few tail curves traced inside an inset were converted to 2θ with the panel's main axis (Supporting Information, Section S5.1). A second, smaller class of error concerns which curves are distinct measurements. In five of the seven curve-count disagreements with the operator, the pipeline kept curves that the protocol excludes, namely calculated curves that escaped the keyword filter and curves repeated inside an inset, and in six of the seven it kept more curves than the operator, so such errors mostly add curves instead of missing measured ones. A stricter keyword filter that also drops inset curves would fix three of these five panels; on the other two, only the caption, which figure analysis did not receive in the production run, names the calculated curves.

**Table 3.** Summary of the accuracy of the extracted curves against three independent references.

| Reference[a] | Metric | n | Pipeline[b] | Control[c] | Interpretation[d] |
|---|---|---|---|---|---|
| Authors' text | Median absolute first-peak difference | 1,175 curves (444 papers) | 0.0956° [0.0812, 0.1181] | 1.52° | 16 times smaller than for other papers' values |
| Authors' text | Curves within 0.3° of the reported first peak | 1,175 curves (444 papers) | 68.9% [65.0, 72.5] | 13.0% | 5.3 times the rate for other papers' values |
| Hand digitization | Median absolute peak-position difference | 118 curves (55 panels) | 0.0672° [0.0500, 0.0872] | 0.176° | 2.6 times smaller than for random pairs |
| Hand digitization | Share of operator peaks matched within 0.5° (mean per curve) | 122 curves (57 panels) | 82.7% | 29.9% | 2.8 times the rate for random pairs |
| Hand digitization | Median profile root-mean-square error (RMSE), 0–100 scale | 125 curves (60 panels) | 6.24 [5.27, 8.63] | 20.6 | 3.3 times smaller than for random pairs |
| Hand digitization | Panels with the same curve count | 60 panels | 88.3% | – | 53 of 60 panels; the 7 others are listed in Supporting Information, Section S5.2 |
| Synthetic ground truth | Median absolute peak-position error | 222 curves per anchor | 0.032° (exact), 0.030° (coarse) | – | Error against exact truth; a coarse anchor adds none |
| Synthetic ground truth | True peaks recovered within 0.5° | 1,206 true peaks (222 curves) per anchor | 98.8% (exact), 98.7% (coarse) | – | Nearly every true peak found with either anchor |
| Synthetic ground truth | Median profile RMSE, 0–100 scale | 222 curves per anchor | 3.36 (exact), 5.29 (coarse) | – | A coarse anchor raises the profile error, not the peak error |

[a] For the authors' text, the reference is the first-peak 2θ value reported in the same paper's text; only curves of the clean admission class are compared. The hand digitization covers 60 panels digitized by one operator who saw no pipeline traces.

[b] Values in brackets are 95% CIs. Hand-digitization medians are weighted to correct for the sampling design, which oversampled grayscale panels.

[c] For the authors' text, each curve is scored against every first-peak value reported in other papers, with statistics over the pooled curve–value pairs; for the hand digitization, each operator curve is scored against a random pipeline curve from a different panel (the random pairing).

[d] Ratios compare the Pipeline and Control values.

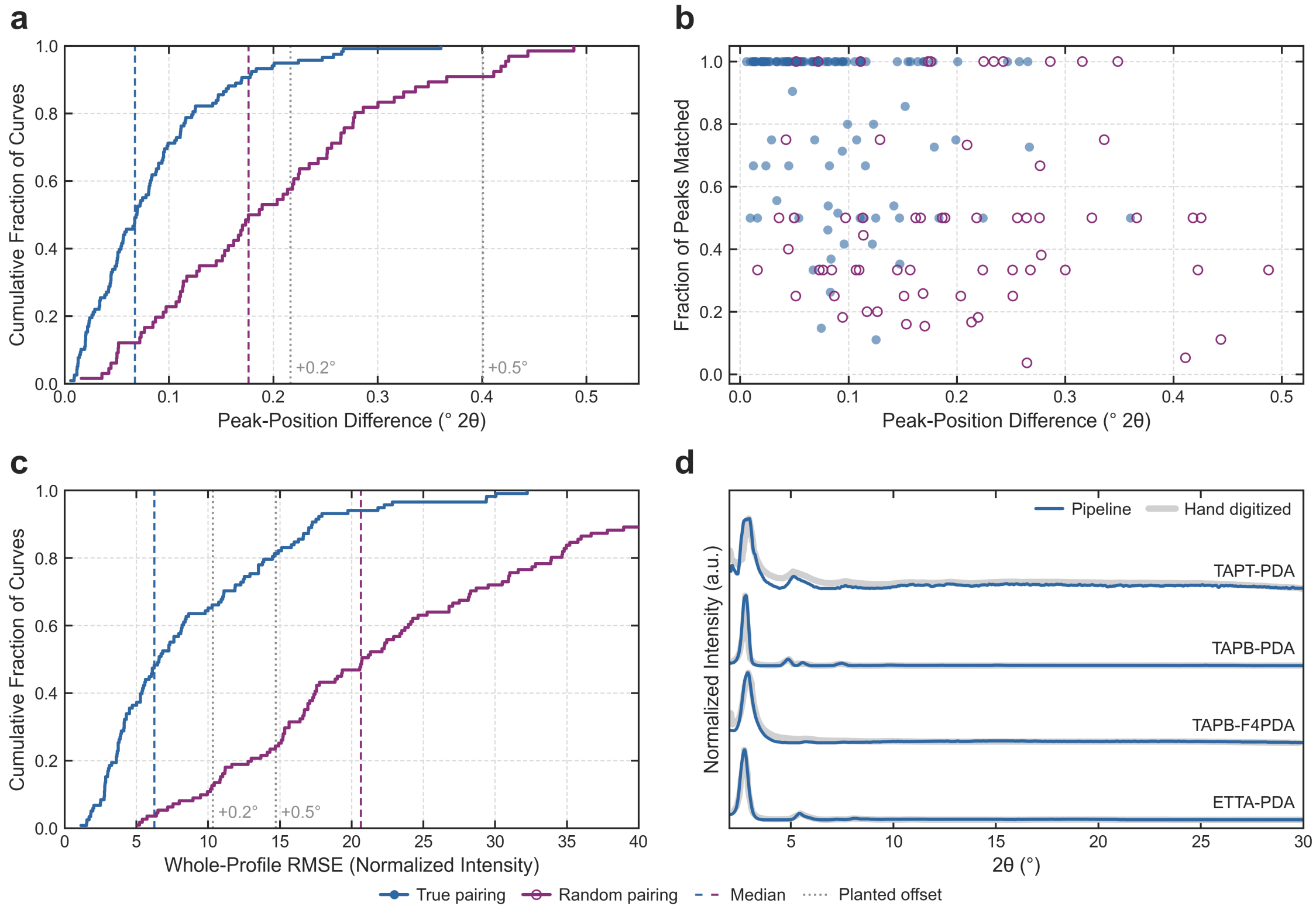

**Figure 7. Comparison of the extracted curves with an independent hand digitization**. (a) Cumulative distributions of the per-curve median difference between peaks matched within 0.5°, for the true (blue, n = 118 curves) and random (purple) pairings, with dotted lines for a planted offset of +0.2° or +0.5° on the hand-digitized curves. (b) Per-curve median difference against the share of operator peaks matched (true, blue, filled; random, purple, open). (c) The whole-profile RMSE on the 0–100 scale (medians, 6.24 true and 20.6 random), with dotted lines for the same planted offsets (10.3 and 14.7). (d) The four curves in Figure 4d of ref. [40] (pipeline, blue line; hand digitization, gray band), with per-curve median peak-position differences of 0.020–0.081°.

## 3.4 Applications of the digitized curves

We ran two analyses on the digitized curves. The first, on synthesis-parameter ladders, uses information that a peak list lacks. The second, on synthesis recipes, points to a next step.

**Synthesis-parameter ladders.** It was observed that some figures show at least three curves of one material made with different reaction times or temperatures, which we call ladders (64 figures, 51 papers). A peak list loses this comparison within one panel. In each ladder, we followed the prominence of the strongest low-angle peak, its height above the neighboring minima, which could be measured in 45 of the 51 papers. The peak became more prominent as the time or temperature increased in 31 of these papers, less prominent in 13, and unchanged in 1 (two-sided sign test, $p = 0.0096$; Supporting Information, Section S6.1). If higher prominence means higher crystallinity, this trend agrees with the common practice of extending the reaction time or raising the temperature to improve crystallinity [2], [3].

**Recipe analyses.** Linking curves to synthesis records also opens questions across papers, such as how the solvent or the reaction temperature shapes crystallinity. Already 1,953 of the accepted curves carry a high-confidence link to a synthesis record. Tests of six recipe variables on these curves estimate that detecting small recipe effects would need about twice as many linked papers (Supporting Information, Section S6.2). Extracting each paper's recipe [6] and linking it to that paper's curves would provide them.

# 4. Concluding remarks

Our pipeline combines an LLM with classical CV. The LLM interprets each figure, and CV measures its geometry and rejects any LLM proposal that fails a pixel check. Applied to 3,954 COF articles, the pipeline produced 11,445 calibrated PXRD curves, which we release in an open database. The curves revealed a consistent trend in peak prominence, a possible marker of crystallinity, within synthesis-parameter ladders, an analysis that a peak list cannot support. In our tests, the LLM was necessary for panel location and curve enumeration. We estimate the median peak-position error of an accepted curve at about 0.03° to 0.07°.

Two insights from this work extend beyond diffraction. First, the tasks requiring the LLM arise from how results are published, in composite figures whose panels overlay the curves of several samples, and not from diffraction itself. We thus expect the same division of labor for UV-Vis spectra [14], chromatograms, and voltammograms published in this form, although no other figure type was tested. Second, the per-task ablation, whose measures do not assume the LLM is correct, offers a reusable way to determine, without ground truth, where a figure-mining workflow needs a vision LLM.

The rates describe this corpus, and a stratified sample of the COF literature is needed to generalize them. The hand digitization, done by one operator, should be repeated to measure the operator's own variability. About 10% of curves place their first peak more than 5° from the text value under every calibration variant, which calls for curating the text-reported peaks. Asking the LLM directly for each curve's 2θ and intensity values [17] and scoring them against the same references would show what pixel-level verification adds. We envision that pairing a vision LLM with pixel-level verification can turn published curves across the chemical literature into calibrated, reusable data. Such data could support agentic AI systems for scientific discovery [41] and the predictive synthesis of metal–organic and covalent organic frameworks [42].

# Acknowledgements

We thank Peter Walther for building the synthesis database used as the text reference.

# Author Contributions

Z.Z. set the direction of the research and supervised the work. S.G. designed and built the pipeline and the website, carried out the analyses, and prepared the manuscript and the Supporting Information.

# Funding

Z.Z. acknowledges funding and extends thanks for the support provided by the Washington University in St. Louis through the Startup Grant.

## Conflicts of Interest

There are no conflicts to declare.

## Data Availability Statement

The 11,445 accepted curves are openly available under a CC BY 4.0 license as files attached to the releases of the GitHub repository https://github.com/AaronG211/pxrd_PEAK. They can also be browsed on the COF-PEAK website https://pxrd-peak.vercel.app and queried through its read-only API at https://pxrd-peak.vercel.app/api/v1. The code is openly available in the same repository under the MIT license.

## References

[1] A. P. Côté, A. I. Benin, N. W. Ockwig, M. O'Keeffe, A. J. Matzger, and O. M. Yaghi, "Porous, Crystalline, Covalent Organic Frameworks," *Science*, vol. 310, no. 5751, pp. 1166–1170, Nov. 2005, doi: 10.1126/science.1120411.

[2] K. Geng *et al.*, "Covalent Organic Frameworks: Design, Synthesis, and Functions," *Chem. Rev.*, vol. 120, no. 16, pp. 8814–8933, Jan. 2020, doi: 10.1021/acs.chemrev.9b00550.

[3] F. Haase and B. V. Lotsch, "Solving the COF Trilemma: Towards Crystalline, Stable and Functional Covalent Organic Frameworks," *Chem. Soc. Rev.*, vol. 49, no. 23, pp. 8469–8500, 2020, doi: 10.1039/d0cs01027h.

[4] M. C. Swain and J. M. Cole, "ChemDataExtractor: A Toolkit for Automated Extraction of Chemical Information from the Scientific Literature," *J. Chem. Inf. Model.*, vol. 56, no. 10, pp. 1894–1904, Oct. 2016, doi: 10.1021/acs.jcim.6b00207.

[5] O. Kononova *et al.*, "Text-Mined Dataset of Inorganic Materials Synthesis Recipes," *Sci. Data*, vol. 6, no. 1, p. 203, Oct. 2019, doi: 10.1038/s41597-019-0224-1.

[6] Z. Zheng, O. Zhang, C. Borgs, J. T. Chayes, and O. M. Yaghi, "ChatGPT Chemistry Assistant for Text Mining and the Prediction of MOF Synthesis," *J. Am. Chem. Soc.*, vol. 145, no. 32, pp. 18048–18062, Aug. 2023, doi: 10.1021/jacs.3c05819.

[7] J. Dagdelen *et al.*, "Structured Information Extraction from Scientific Text with Large Language Models," *Nat. Commun.*, vol. 15, no. 1, p. 1418, Feb. 2024, doi: 10.1038/s41467-024-45563-x.

[8] J. Maschita, T. Banerjee, and B. V. Lotsch, "Direct and Linker-Exchange Alcohol-Assisted Hydrothermal Synthesis of Imide-Linked Covalent Organic Frameworks," *Chem. Mater.*, vol. 34, no. 5, pp. 2249–2258, Feb. 2022, doi: 10.1021/acs.chemmater.1c04051.

[9] A. Rohatgi, *WebPlotDigitizer*. Automeris. [Online]. Available: https://automeris.io

[10] F. Marin, A. Rohatgi, and S. Charlot, "WebPlotDigitizer, a Polyvalent and Free Software to Extract Spectra from Old Astronomical Publications: Application to Ultraviolet Spectropolarimetry," in *SF2A-2017: Proceedings of the Annual Meeting of the French Society of*

*Astronomy and Astrophysics*, 2017, pp. 113–117. [Online]. Available: https://sf2a.eu/proceedings/2017/2017sf2a.conf..0113M.pdf

[11] D. Drevon, S. R. Fursa, and A. L. Malcolm, "Intercoder Reliability and Validity of WebPlotDigitizer in Extracting Graphed Data," *Behav. Modif.*, vol. 41, no. 2, pp. 323–339, 2017, doi: 10.1177/0145445516673998.

[12] M. Mitchell *et al.*, *Engauge Digitizer*. (Jul. 12, 2020). Zenodo. doi: 10.5281/ZENODO.3941227.

[13] J. Lal, A. Mitkari, M. Bhosale, and D. Doermann, "LineFormer: Line Chart Data Extraction Using Instance Segmentation," in *Document Analysis and Recognition – ICDAR 2023*, Cham: Springer Nature Switzerland, 2023, pp. 387–400. doi: 10.1007/978-3-031-41734-4_24.

[14] W. Jiang, K. Li, T. Spreadbury, E. Schwenker, O. Cossairt, and M. K. Y. Chan, "Plot2Spectra: An Automatic Spectra Extraction Tool," *Digit. Discov.*, vol. 1, no. 5, pp. 719–731, 2022, doi: 10.1039/d1dd00036e.

[15] E. Schwenker, W. Jiang, T. Spreadbury, N. Ferrier, O. Cossairt, and M. K. Y. Chan, "EXSCLAIM!: Harnessing Materials Science Literature for Self-Labeled Microscopy Datasets," *Patterns*, vol. 4, no. 11, p. 100843, Nov. 2023, doi: 10.1016/j.patter.2023.100843.

[16] Z. Zheng *et al.*, "Image and Data Mining in Reticular Chemistry Powered by GPT-4V," *Digit. Discov.*, vol. 3, no. 3, pp. 491–501, 2024, doi: 10.1039/d3dd00239j.

[17] M. P. Polak and D. Morgan, "Leveraging Vision Capabilities of Multimodal LLMs for Automated Data Extraction from Plots." arXiv, 2025. doi: 10.48550/ARXIV.2503.12326.

[18] J. Luo, Z. Li, J. Wang, and C.-Y. Lin, "ChartOCR: Data Extraction from Charts Images via a Deep Hybrid Framework," in *2021 IEEE Winter Conference on Applications of Computer Vision (WACV)*, IEEE, Jan. 2021, pp. 1916–1924. doi: 10.1109/wacv48630.2021.00196.

[19] N. Alampara *et al.*, "Probing the Limitations of Multimodal Language Models for Chemistry and Materials Research," *Nat. Comput. Sci.*, vol. 5, no. 10, pp. 952–961, Aug. 2025, doi: 10.1038/s43588-025-00836-3.

[20] G. Bradski, "The OpenCV Library," *Dr Dobb's J. Softw. Tools*, vol. 25, no. 11, pp. 120–125, Nov. 2000.

[21] P. Rahmanzadehgervi, L. Bolton, M. R. Taesiri, and A. T. Nguyen, "Vision Language Models Are Blind," in *Computer Vision – ACCV 2024*, Singapore: Springer Nature Singapore, 2025, pp. 293–309. doi: 10.1007/978-981-96-0917-8_17.

[22] Y. Du *et al.*, "PP-OCR: A Practical Ultra Lightweight OCR System." arXiv, 2020. doi: 10.48550/ARXIV.2009.09941.

[23] X. Lei, Z. Yang, X. Chen, P. Li, and Y. Liu, "Scaffolding Coordinates to Promote Vision-Language Coordination in Large Multi-Modal Models," in *Proceedings of the 31st*

*International Conference on Computational Linguistics (COLING 2025)*, 2025, pp. 2886–2903. [Online]. Available: https://aclanthology.org/2025.coling-main.195/

[24] P. H. C. Eilers and H. F. M. Boelens, “Baseline Correction with Asymmetric Least Squares Smoothing,” Leiden University Medical Centre, Leiden, The Netherlands, 2005.

[25] A. Savitzky and M. J. E. Golay, “Smoothing and Differentiation of Data by Simplified Least Squares Procedures,” *Anal. Chem.*, vol. 36, no. 8, pp. 1627–1639, Jul. 1964, doi: 10.1021/ac60214a047.

[26] M. Everingham, L. Van Gool, C. K. I. Williams, J. Winn, and A. Zisserman, “The Pascal Visual Object Classes (VOC) Challenge,” *Int. J. Comput. Vis.*, vol. 88, no. 2, pp. 303–338, 2010, doi: 10.1007/s11263-009-0275-4.

[27] P. Virtanen *et al.*, “SciPy 1.0: Fundamental Algorithms for Scientific Computing in Python,” *Nat. Methods*, vol. 17, no. 3, pp. 261–272, Feb. 2020, doi: 10.1038/s41592-019-0686-2.

[28] M. Lipsitch, E. Tchetgen Tchetgen, and T. Cohen, “Negative Controls,” *Epidemiology*, vol. 21, no. 3, pp. 383–388, May 2010, doi: 10.1097/ede.0b013e3181d61eeb.

[29] B. Efron, “Bootstrap Methods: Another Look at the Jackknife,” *Ann. Stat.*, vol. 7, no. 1, pp. 1–26, Jan. 1979, doi: 10.1214/aos/1176344552.

[30] C. A. Field and A. H. Welsh, “Bootstrapping Clustered Data,” *J. R. Stat. Soc. Ser. B Stat. Methodol.*, vol. 69, no. 3, pp. 369–390, May 2007, doi: 10.1111/j.1467-9868.2007.00593.x.

[31] B. A. Nosek, C. R. Ebersole, A. C. DeHaven, and D. T. Mellor, “The Preregistration Revolution,” *Proc. Natl. Acad. Sci.*, vol. 115, no. 11, pp. 2600–2606, Mar. 2018, doi: 10.1073/pnas.1708274114.

[32] A. G. Seco de Herrera, R. Schaer, S. Bromuri, and H. Müller, “Overview of the ImageCLEF 2016 Medical Task,” in *CLEF 2016 Working Notes*, in CEUR Workshop Proceedings, vol. 1609. 2016, pp. 219–232. [Online]. Available: https://ceur-ws.org/Vol-1609/16090219.pdf

[33] S. Tsutsui and D. J. Crandall, “A Data Driven Approach for Compound Figure Separation Using Convolutional Neural Networks,” in *2017 14th IAPR International Conference on Document Analysis and Recognition (ICDAR)*, IEEE, Nov. 2017, pp. 533–540. doi: 10.1109/icdar.2017.93.

[34] W. Jiang, E. Schwenker, T. Spreadbury, N. Ferrier, M. K. Y. Chan, and O. Cossairt, “A Two-Stage Framework for Compound Figure Separation,” in *2021 IEEE International Conference on Image Processing (ICIP)*, IEEE, Sep. 2021, pp. 1204–1208. doi: 10.1109/icip42928.2021.9506171.

[35] Z. Wang *et al.*, “CharXiv: Charting Gaps in Realistic Chart Understanding in Multimodal LLMs,” in *Advances in Neural Information Processing Systems 37*, Neural Information Processing Systems Foundation, Inc. (NeurIPS), 2024, pp. 113569–113697. doi: 10.52202/079017-3609.

[36] D. Bertillo, G. Melchiorri, and P. Merialdo, “PaperUnPlot: Benchmarking Chart-to-Table in the Wild,” in *4th International Workshop on Tabular Data Analysis (TaDA), co-located with VLDB 2026*, 2026. [Online]. Available: https://tabular-data-analysis.github.io/tada2026/papers/TaDA26_13.pdf

[37] L. Foppiano, G. Lambard, T. Amagasa, and M. Ishii, “Mining Experimental Data from Materials Science Literature with Large Language Models: An Evaluation Study,” *Sci. Technol. Adv. Mater. Methods*, vol. 4, no. 1, p. 2356506, Jul. 2024, doi: 10.1080/27660400.2024.2356506.

[38] H. W. Kuhn, “The Hungarian Method for the Assignment Problem,” *Nav. Res. Logist. Q.*, vol. 2, no. 1–2, pp. 83–97, Mar. 1955, doi: 10.1002/nav.3800020109.

[39] J. D. Hunter, “Matplotlib: A 2D Graphics Environment,” *Comput. Sci. Eng.*, vol. 9, no. 3, pp. 90–95, 2007, doi: 10.1109/mcse.2007.55.

[40] L. D. Tran *et al.*, “Conversion of Soluble Polyimines to Covalent Organic Framework Films and Composites,” *J. Am. Chem. Soc.*, vol. 147, no. 28, pp. 24413–24421, Jul. 2025, doi: 10.1021/jacs.5c03079.

[41] B. Feng *et al.*, “Agentic AI Systems for Scientific Discovery,” *Nat. Rev. Methods Primers*, vol. 6, no. 1, p. 81, Oct. 2026, doi: 10.1038/s43586-026-00534-y.

[42] B. Feng, R. A. Coyle, S. C. Pal, P. Walther, S. Park, and Z. Zheng, “Predictive Synthesis of MOFs and COFs.” arXiv, 2026. doi: 10.48550/ARXIV.2610.12739.

# Supporting Information

# Structured Extraction of Diffraction Data from Covalent Organic Frameworks

Shenghan Gao[1,2] and Zhiling Zheng[3,*]

[1]Department of Computer Science and Engineering, Washington University, St. Louis, Missouri 63130, United States

[2]Department of Statistics and Data Science, Washington University, St. Louis, Missouri 63130, United States

[3]Department of Chemistry, Washington University, St. Louis, Missouri 63130, United States

*Corresponding author. Email: z.z@wustl.edu

**Contents**

# Section S1. General Information

## S1.1. Model, Software and Code

**Terms.** The Supporting Information uses the terms that main-text Sections 2.2 to 2.5 define, including figure attempt, stored figure, accepted curve, plot box, plot frame, anchor and ink snapping.

**Model and software.** All four large language model (LLM) calls went to GPT-5.5 (OpenAI; dated version gpt-5.5-2026-04-23 in the run's settings) through the Responses API. The calls used structured outputs and no sampling parameter, and each request was answered once and cached. The software of Table S1, including PP-OCR [1] and SciPy [2], ran on one laptop (Apple MacBook Pro, M1 Pro chip, 16 GB of memory). Optical character recognition (OCR) ran on the CPU of this laptop.

**Table S1.** Software of the production run, at the versions installed in the project environment before and after the run unless marked[a].

| Software | Version | Used for |
|---|---|---|
| Python | 3.11.15 | All code |
| PyMuPDF | 1.27.2.2 | Rendering pages and panels; page text |
| RapidOCR (rapidocr-onnxruntime), with its PP-OCR models; ONNX Runtime | 1.4.4; 1.24.4 | OCR of tick labels and of text inside the plot |
| OpenCV (opencv-contrib-python) | 4.10.0.84 | Plot frame, tick marks, color masks, ink |
| NumPy; SciPy | 2.4.4; 1.17.1 | Line fits and interpolation; background, smoothing and peak finding |
| Pillow | 12.2.0 | Resizing and PNG encoding of the images sent to the LLM |
| OpenAI Python library | 2.32.0 | LLM calls |

**Code and data.** The code is hosted at https://github.com/AaronG211/pxrd_PEAK under the MIT license. Its releases carry the 11,445 accepted curves under the CC BY 4.0 license (Section S7.1).

**Run record.** The production run took place from 13 June to 2 July 2026. The code changed during the run, and the only change known to affect released curves is the keyword filter, added after the first 87 figure attempts (Section S2.2). The code was placed under version control after the run, and the released pipeline code is the first commit plus one later change. That change only records which mask each trace used, and Section S2.8 lists the known differences from the run. Rendering, OCR, computer vision (CV) and post-processing are deterministic, and the LLM's answers are part of the data.

## S1.2. Corpus

**Provenance.** The Web of Science export [3] held 4,635 records with 4,633 distinct digital object identifiers (DOIs). PDFs were obtained for 4,286 of the 4,633 articles (92.5%). Processing in DOI order placed RSC (prefix 10.1039) last among the large publishers, and the provider quota stopped the run after the first 3,954 PDFs. The corpus is the 2,554 articles of the run (Table S2), and 2,548 of them finished with at least one located panel. The other 6 were still in progress at the stop, 2 of them before any figure attempt.

**Table S2.** Downloaded articles by their outcome in the production run, with both text measures of X-ray diffraction (XRD) mentions counted in the article PDF, not in its supporting information.

| Group | Articles | Pages | Never mention XRD[a] (%) | With an XRD caption[b] (%) |
|---|---|---|---|---|
| In the run | 2,554 | 24,086 | 1.0 | 93.0 |
| Processed, no panel found | 1,400 | 13,403 | 44.3 | 2.9 |
| Never processed (run stopped) | 332 | 3,042 | 13.6 | 63.0 |
| Not in the run (two rows above) | 1,732 | 16,445 | 38.4 | 14.4 |

[a] The text contains none of "XRD", "PXRD", "X-ray (powder) diffraction", "diffractogram", "powder diffraction" or "diffraction pattern", in any letter case.

[b] A line that starts like a figure caption ("Fig. 3", "Figure S2", "Scheme 1") and has one of these terms in it or in the next two lines.

**Retention.** Panel location kept 2,376 of the 2,417 processed articles with an XRD caption (98.3%). Retention was 97.9% to 98.7% of captioned articles at each of RSC, ACS, Wiley and Elsevier. RSC articles (DOI prefix 10.1039) are under-represented mainly because of the stop. Only 218 of the 798 downloaded RSC articles are in the run, and 325 of the 332 never-processed articles are from RSC.

## S1.3. Reference Data from the Authors' Text

**Synthesis database.** Peter Walther built the synthesis database from 1,161 articles with a separate, AI-based workflow. It holds 2,933 records in 211 columns, and each record is one covalent organic framework (COF) in one paper. For 2,745 records, the workflow also read the supporting information. The synthesis database is not publicly released. This work uses its peak lists, names, structure class, building-block connectivity and six recipe variables.

**Peak lists.** Of the 2,933 records, 1,464 carry a peak list, which is free text such as "5.6; 9.7; 11.5". The parser turns each list into 2θ values with the seven rules of Table S3, applied in order. It returns values for 1,459 of the 1,464 lists. The first reported peak of a material is the smallest value in its list.

**Table S3.** Rules of the peak-list parser, applied in order to the free text of each peak list.

| Rule | Field content | Result |
|---|---|---|
| (i) | Empty, or "no report" | No list |
| (ii) | Mentions "broad" and has no number followed by a semicolon or comma | No list |
| (iii) | Miller indices such as "(001)" | Removed |
| (iv) | References such as "Table S1" | Removed |
| (v) | Whole ranges such as "20–25" | Removed |
| (vi) | Bounds such as ">20" | Removed |
| (vii) | Each remaining number between 0.5 and 90 (with "~" and words ignored) | Kept as a 2θ value in degrees |

**Name matching.** Only curves of the clean admission class (Section S7.2) are linked, each by name to a record of its own paper. The match compares the curve's legend label and material name, both read by the LLM, with the record's COF name, alias and sample label. Names are reduced to lowercase letters and digits, and names under three characters are ignored. Two names match when one contains the other.

**Tiers.** A curve falls into tier A when a record matches, and an exact match is preferred over the longest containment. A tie between records is left ambiguous and unused. A curve falls into tier B when no record matches in a paper with one record, and into the unused tier C when the paper has several records. Tiers A and B give "high-confidence" links. The 2,760 clean curves of Table S4 gave 1,451 tier-A, 135 ambiguous, 502 tier-B and 672 tier-C links.

**Independence and errors.** The database holds no curve data, and our pipeline never reads it, so agreement is not built in. The peak lists of the database contain extraction errors [4]. For CORN-COF-1, the article names reflections at 4.5° and 5.1°, but the database stores only 5.1°.

**Scan check.** For 377 of the 668 materials of Table S4, a rule-based scan of the article PDFs matched 1,258 of 1,392 stored values (90.4%) within 0.05°. This share is not an error rate, because the scan itself misses values. Wrong links between curves and records also remain possible.

**Table S4.** From accepted curves to curves compared with the authors' text, where each row keeps the curves of the row above that meet one more condition.

| Condition | Curves | Papers |
|---|---|---|
| Accepted curve | 11,445 | 2,370 |
| Paper has a record in the synthesis database | 3,809 | 847 |
| Curve is in the clean class (Section S7.2) | 2,760 | 734 |
| High-confidence link (to 1,037 distinct records) | 1,953 | 610 |
| Linked record has a parsable peak list (668 records) | 1,202 | 447 |
| Curve has a first peak of status ok or low_confidence (Section S7.1) | 1,175 | 444 |

**Figure 1b,c.** Main-text Figure 1b,c uses the 668 materials of Table S4. For material $m$, the share of its plotted 2θ range above its highest reported angle is

$$h_m = \frac{\bar{a}_m^{\max} - t_m^{\max}}{\bar{a}_m^{\max} - \bar{a}_m^{\min}}.$$

Here $\bar{a}_m^{\min}$ and $\bar{a}_m^{\max}$ are the ends of the 2θ range of its linked curves, averaged over the curves, and $t_m^{\max}$ is its highest reported angle.

**Estimates.** Peak lists hold one value for 107 materials (16.0%), with a 95% confidence interval (CI; Section S1.4) of 11.4% to 21.0%. They hold at most three values for 321 materials (48.1%; 95% CI, 42.7% to 53.4%). The median $h_m$ is 50.6% (95% CI, 42.8% to 59.3%), or 50.3% (95% CI, 45.5% to 56.3%) with one material per paper (447 materials).

## S1.4. Statistics and Analysis Plans

**Bootstrap.** Every 95% CI comes from a cluster bootstrap [5], [6], which recomputes a statistic on many resampled copies of the data. Each draw resamples whole papers, or whole panels for the hand digitization. The CI spans the 2.5th to 97.5th percentiles of the resampled values. The analyses use between 2,000 and 10,000 draws, each with a fixed seed.

**Written plans.** For five analyses, the plan was recorded before the analysis script first ran. The plan for anchor tracing (Section S4.5) fixed its cohort, eight baselines, 60-point anchors, median first-peak difference, coverage and decision rule. The plan for the hand digitization (Section S5.2) fixed 60 panels in six strata and a seed from the plan. It also fixed a blinded operator, five outcomes and a re-digitization of 15 panels.

**Application plans.** The plan for the ladder robustness check (Section S6.1) fixed seven settings (2,880 combinations per measure), two measures and a sign test. The plan for the recipe tests (Section S6.2) fixed 24 tests, their statistic, null, decision rule and controls. The plan for their power analysis (Section S6.2) fixed effect sizes, three patterns, critical values from 20,000 shuffles and 80% power.

**Departures.** Only departures that change how a result reads are given here. In anchor tracing, matched processing replaced the plan's comparison, and coverage became the main result. In the hand digitization, the re-digitization was not made, so random pairing is the control. Exclusions were also supplied on 38 panels.

**Post hoc choices.** The two ladder measures were chosen after the first prominence result and after an earlier, unrecorded sweep had been seen. The recipe rerun with chemically named categories only is post hoc. The power plan was written after the recipe results were known, but before any power value.

**Other analyses.** The layout measures (Sections S4.2 and S4.3) had no written plan. The tick-reading plan (Section S4.4) is not counted. Its file is rewritten on each run, so the plan cannot be shown to predate the first result.

# Section S2. Pipeline Implementation

## S2.1. Rendering and Panel Location

**Rendering.** The pipeline renders each PDF page at 150 dpi to locate its panels (Section S3.2). Each located panel was then padded and rendered again at a median of 555 dpi (limits in Section S2.9).

**Pixel width.** The 2θ width of one pixel is the slope $m$ of the calibration line (Section S2.4). Its median over the 4,664 stored figures is 0.0337°, and it exceeds 0.02° in 95.1% of them (Figure S1a). Placing a peak within 0.1° therefore needs a trace accurate to about three pixels.

**Anchor spacing.** The 11,445 accepted curves span a median of 34.9° (Figure S1b), or 1,055 pixels. With a median of 49 points (Section S2.5), an anchor spaces its points about 22 pixels apart, far coarser than these three pixels.

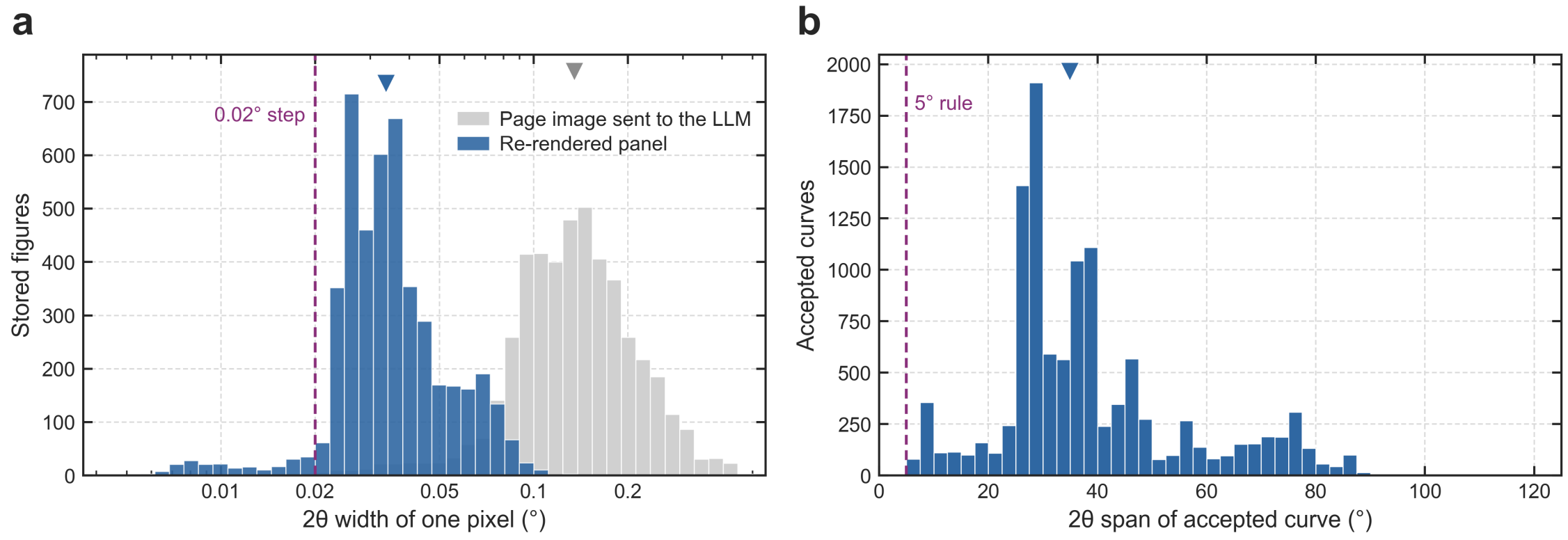

**Figure S1. Pixel resolution of the panels and 2θ span of the curves.** (a) 2θ width of one pixel on the re-rendered panel (blue) and the page image (gray). (b) 2θ span of accepted curves. Re-rendering makes a pixel four times narrower (median 0.034° against 0.135°).

## S2.2. Figure Analysis and the Keyword Filter

**Plot box and reframing.** Figure analysis returns, for each panel, the LLM's plot box, tick values and list of curves (Section S3.3). When the plot box nearly touches one image border or leaves a wide empty margin, the panel is cut again with the cut side extended. The new cut is analyzed again, for at most two rounds (thresholds in the released code).

**Keyword filter.** The keyword filter keeps simulated, refined and other model-derived curves out of tracing (Table S5). On panels judged to be diffraction patterns, the LLM listed 18,726 curves, 892 of them stick patterns. The released filter marks 4,952 of the other 17,834 as model-derived.

**Production filter.** The production filter was absent for the first 87 figure attempts and lacked the words marked †, such as "stacking". The 204 model-derived curves accepted as a result are released as model_derived (Section S7.2). Of these, 98 came from those first attempts and 106 carried a † word.

**Table S5.** Words of the keyword filter in the released code, matched regardless of case.[a]

| Text searched | Match rule | Words |
|---|---|---|
| Sample state | Whole text equals the token | sim, calc, calcd, diff, ycal, obs-calc |
| Sample state | Text contains the stem | simulat, calculat,[b] pawley, pawly, rietveld, refine, difference, residual, theoret, theory, predicted, modeled†, model-derived, dft-derived†, stacking†, eclipsed†, staggered† |
| Label and material name, only when the sample state is empty | Text contains the marker | simulat, (sim, -sim, calculat, pawley, rietveld, refined, difference, ycal, stacking†, eclipsed, staggered |

[a] Words marked † were absent from the production run's list for that row. [b] The stem "calculat", not "calc", leaves the measured state "calcined" unmarked.

## S2.3. Plot Frame and Tick Marks

**Plot frame.** CV finds the plot frame, the rectangle bounded by the axes, from candidate bottom axes, which are long horizontal runs of dark pixels (Section S2.9). Each candidate frame is scored by the tick marks on its bottom line, the numbers that OCR reads below it, its share of ink and its width. The search starts near the LLM's plot box, and two more frames are tried in turn if calibration fails.

**Tick marks.** Tick marks are pixel columns that are mostly dark within narrow bands just below and just above the axis line. They are fitted to an evenly spaced grid, which fills in marks missed inside wider gaps. The other thresholds and weights of the frame search are those of the released code.

## S2.4. Axis Calibration

**Two fits.** Axis calibration fits the calibration line $2\theta = m\,x + q$ by least squares to pairs of tick position and 2θ value. It makes two such fits, the OCR fit and the LLM fit, and compares them. Up to two outlier pairs with large residuals are dropped from a fit (Section S2.9). The OCR fit places each tick label that OCR reads below the axis line at the center of its box. The LLM fit places the LLM's tick values on the CV tick marks.

**Placing the LLM's values.** When marks and values are equal in number, the values go on the marks in order. Otherwise the evenly spaced values slide along the marks, which may include unlabeled minor marks. Every placement with a small root-mean-square error (RMSE) is kept (Section S2.9).

**Choosing a placement.** When several placements remain, the first of three rules that applies picks one (thresholds in Section S2.9). Rule (i) takes the placement closest to an OCR fit that passed its checks. Rule (ii) takes a placement close to every OCR label when the next best misses some label widely. Rule (iii) takes the only placement whose labels span most of the marks. If no rule applies, no LLM fit is made.

**Fit checks.** Each fit must then pass checks for a plausible axis (Section S2.9). Over the range of its supporting points, it must rise from left to right, with its 2θ values and span within set limits. Its RMSE limit is stricter for the LLM fit than for the OCR fit.

**Consensus.** Two passing fits reach consensus when their 2θ values differ by at most 0.5° at evenly spaced columns across the plot frame. The fit with the smaller RMSE is then used (Table S6). When only one fit passes, the lone-fit rule decides whether to accept it. The rule asks for at least three ticks and an RMSE of at most 0.15°.

**Dependent fits.** Rule (i) makes the two fits dependent, and it decided 2,541 of the 3,825 consensus calibrations. Without rule (i), 1,827 of them would have missed consensus. Of these, 1,784 would still have passed the lone-fit rule on the OCR fit alone.

**Axis arbitration.** For a figure attempt without an accepted fit, axis arbitration has the LLM reread the tick values from a threefold enlarged strip around the axis line (Section S3.5). The reread values need a single placement on the CV tick marks, and the new fit needs a plausible axis and a small RMSE (Section S2.9). Arbitration ran on 558 figure attempts and recalibrated 165 of them. In the others, the reread values could not be placed on the tick marks (301), the new fit failed (5), or the call returned fewer than two values or no answer (87).

**Table S6.** Calibration outcome of the 5,295 figure attempts of the production run (main text Figure 3d).

| Status | Meaning | Figure attempts | Share |
|---|---|---:|---:|
| Consensus | Both fits passed and agreed within 0.5° | 3,825 | 72.2% |
| Lone fit | One fit passed and met the lone-fit rule (OCR fit 624, LLM fit 116) | 740 | 14.0% |
| Arbitrated | Recalibrated by axis arbitration | 165 | 3.1% |
| Failed | No fit passed its checks; arbitration did not help | 193 | 3.6% |

| Status | Meaning | Figure attempts | Share |
| --- | --- | --- | --- |
| Disagreement | Both fits passed but disagreed; arbitration did not help | 160 | 3.0% |
| Weak lone fit | One fit passed but failed the lone-fit rule; arbitration did not help | 40 | 0.8% |
| Not reached | Rejected before calibration (Section S2.7) | 172 | 3.2% |

## S2.5. Anchors, Ink Snapping and Acceptance Gates

**Anchor.** Ink snapping turns each anchor, the LLM's rough path of one curve, into a trace on the ink. The anchor height $A(x)$ joins the anchor points (Section S3.4) by straight lines and is never extrapolated beyond them. An anchor needs a minimum number of points (Section S2.9), and the run's anchors held a median of 49.

**Ink.** Ink is the set of pixels that may belong to a drawn line. A pixel is ink when it is dark for its frame or strongly colored (Section S2.9). Text found by OCR is erased, and axis lines are kept.

**Masks.** The three masks, tried in this order, set the candidate ink $M$ and the search half-width $w$ in pixels for a plot frame $H$ pixels high. The search half-width is how far from the anchor height ink is sought. Section S2.8 gives how the run chose among the masks.

**(i) Color mask.** The color mask keeps only ink of the curve's color, using a fixed range of hue, saturation and value for each color word. Of the 11,880 curve attempts, 643 (5.4%) carried an unlisted word and had no color mask. Of the 11,445 accepted curves, 85.3% were traced on this mask.

**(ii) Ink mask.** The ink mask keeps all ink, with the same $w$ as the color mask. The run traced 14.2% of the accepted curves on it.

**(iii) Wide mask.** The wide mask also keeps all ink, with a larger $w$. The run traced the remaining 0.5% of the accepted curves on it.

**Snapping.** In each column where $A(x)$ is defined, the trace $\hat{y}(x)$ is the weighted mean height of the candidate ink in the search window $S(x)$ (Figure S2a–c). Ink close to the anchor height weighs most.

$$S(x) = \{y: (x, y) \in M,\ |y - \lfloor A(x) \rceil| \le w\},$$

$$\hat{y}(x) = \frac{\sum_{y \in S(x)} \omega(y)\, y}{\sum_{y \in S(x)} \omega(y)}, \qquad \omega(y) = \frac{1}{1 + \left(\frac{|y - A(x)|}{\max(1,\ w/2)}\right)^2}.$$

**Unsnapped columns.** A column with an empty $S(x)$, for example where a label covers the curve, is unsnapped and keeps the anchor height. Its height thus comes from the LLM, not the pixels. Trace heights far from their running median are then replaced by that median (Section S2.9). Columns before the first or after the last snapped column are deleted.

**Snap rate.** The snap rate $s$ is the share of the remaining columns that were snapped. Unsnapped columns make up 1.9% of an accepted curve's columns on average and occur in 42.7% of accepted curves.

**Scores.** The acceptance gates test each trace with four scores, of which the snap rate is the first. Let $X$ be the trace columns and $d(x)$ the distance from $\hat{y}(x)$ to the nearest pixel of $M$ in column $x$. Section S2.9 gives the tolerance $\tau$ and the band half-width $b$ in pixels. Precision $p$ is the share of trace columns on ink. Recall $r$, as the production run computed it, is that share among the trace columns $B_t$ with ink within $b$ of the trace.

$$p = \frac{|\{x \in X: d(x) \leq \tau\}|}{|X|}, \qquad r = \frac{|\{x \in B_t: d(x) \leq \tau\}|}{|B_t|}.$$

The released code instead counts the ink within $b$ of the anchor, including abandoned columns (Section S2.8).

**Confidence.** A fourth score, the confidence $c$, adds the agreement $\delta$ between trace and anchor.

$$c = 0.30\,s + 0.30\,p + 0.25\,r + 0.15\max\left(0,\ 1 - \frac{\delta}{0.25}\right), \qquad \delta = \frac{1}{H}\sqrt{\frac{1}{|X|}\sum_{x \in X}\left(A(x) - \hat{y}(x)\right)^2}.$$

**Gates.** A trace must pass all four gates, $s \geq 0.55$, $p \geq 0.70$, $r \geq 0.60$ and $c \geq 0.55$. Under the production recall, only precision and snap rate could reject a trace. Every trace has $r \geq p$, so passing precision passes recall. Likewise, any trace with $s \geq 0.55$ and $p \geq 0.70$ already has $c \geq 0.55$. All 170 traces that the gates rejected had failed precision.

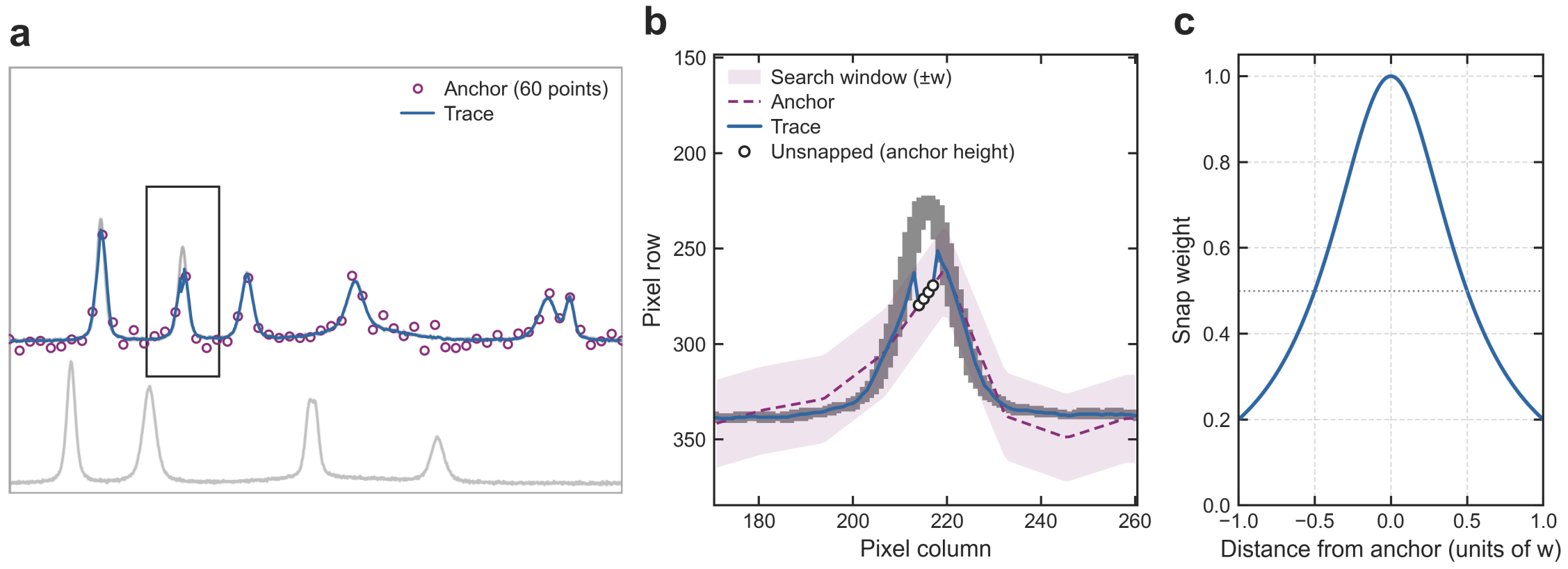

**Figure S2. Ink snapping on a synthetic panel (Section S5.3).** (a) Trace from a coarse 60-point anchor ($w$ = 23 px). (b) The boxed peak; ink in gray. The trace follows the ink, except in four columns with no ink inside the search window, which keep the anchor height. (c) Snap weight $\omega$.

## S2.6. Post-processing

**Coordinates.** Post-processing converts each trace into 2θ and intensity and then applies the steps of Table S7. Trace column $x$ becomes $2\theta = m\,(x_0 + x) + q$, with $x_0$ the left edge of the plot frame. The intensity of the column is then $I = H - 1 - \hat{y}(x)$.

**Background.** The background step fits a smooth baseline $z$ that stays under the peaks. The baseline minimizes the asymmetric least-squares (AsLS) objective $\sum_i \phi_i\,(I_i - z_i)^2 + \lambda \sum_i (z_{i-1} - 2z_i + z_{i+1})^2$ over the trace columns $i$ [7]. In it, $\lambda$ sets the smoothness, and the asymmetry $\kappa$ gives $\phi_i = \kappa$ where $I_i > z_i$ and $1 - \kappa$ elsewhere.

**Stored values.** Stored intensities run from −15.1 to 114.5, because the run did not clip after smoothing. Of the 21,859,711 stored points, 1.4% lie below 0 and 0.03% above 100. Over the 11,445 accepted curves, the 2θ uncertainty $u$ has a median of 0.048° (10th to 90th percentile, 0.030° to 0.094°).

**Table S7.** Post-processing steps of the production run, in the order applied.

| Step | Setting |
|---|---|
| Calibrated window[a] | Trace clipped to the tick positions ± 1.25 median gaps, within the plot frame and 0.2° to 130° (the frame alone with fewer than two positions); rejected if less than 30% of the trace lies inside |
| Span rule | Rejected if the clipped trace spans less than 5° |
| Background (AsLS) | $\lambda = 10^7$, $\kappa = 0.01$, 10 iterations; 5th percentile shifted to 0; negative values set to 0 |
| Normalization | Largest value scaled to 100 |
| Resampling | Linear, 0.02° step from the trace's own first point |
| Smoothing | Savitzky–Golay [8], 11 points (0.20°), cubic; no clipping afterward |
| 2θ uncertainty | $u$ = fit RMSE + $\lvert m \rvert$, one value per figure |

[a] The constants of the calibrated window are those of the released code. The production run did not record which positions it pooled (Section S2.8).

## S2.7. Figure Outcomes, Rejections and Runtime

**Rejections.** The 172 figure attempts rejected before calibration (Table S6) were non-PXRD panels, undetected frames, provider quota errors and panels without a traceable curve (main text Figure 5b). A measured curve without an anchor counts as a rejected curve attempt ("No trace returned" in main text Figure 5c). The 66 "No curve accepted" attempts of Figure 5b carry the recorded reason "all series failed fidelity verification", whatever caused their curves to fail.

**Runtime.** A figure attempt took a median of 57.6 s (10th to 90th percentile, 30.4 to 122.9 s), or 106.1 h in all. These times exclude page rendering and panel location.

**Span rule.** The span rule (Section S2.6) caused 258 of the 435 curve rejections (59.3%). Of these, 155 (in 40 figures) came from close-up panels whose 2θ axis spans only a few degrees, which were rejected by design. In the inspected trace overlays, most of the rest were slivers, traces that follow only a short fragment of a curve running across the whole axis. None of the inspected traces came from an anchor on a wrong panel.

## S2.8. Production Run and Released Code

**Table S8.** The six known differences between the production run and the released code.

| Step | Production run | Released code | Effect on the released data |
|---|---|---|---|
| Keyword filter (Section S2.2) | Absent for the first 87 figure attempts; lacked the † words | Full list of Table S5 | 204 model-derived curves accepted; released as model_derived |
| Recall gate (Section S2.5) | Ink within $b$ of the trace; untraced columns skipped | Ink within $b$ of the anchor; abandoned columns count | Only snap rate and precision could reject a trace |
| Choice among the masks (Section S2.5) | Highest $0.4\,s + 0.4\,p + 0.2\,r$ among the masks tried | First mask whose trace passes the gates | Not recorded, so not quantified |
| Pairing anchors with curves (Section S3.4) | By list position | By the curve's number from the top | Differ in 469 of 4,713 reproduced anchor calls; no curve lost in the run |
| Calibrated window (Section S2.6) | Tick marks and OCR labels; exact rule not recorded | Pooled positions of every source that supports the accepted fit | Would clip at least 1,547 of 11,095[a] accepted curves (median 2.8°); tick marks alone clip none |
| Smoothing (Section S2.6) | No step after Savitzky–Golay | Negative values set to 0; maximum rescaled to 100 | Some stored intensities fall outside 0 to 100 |

[a] The 11,095 curves are the 11,445 accepted curves less the 350 on arbitrated axes (Section S7.2).

## S2.9. Key Constants

Table S9 gathers the numeric thresholds and settings of Sections S2.1 to S2.5. Post-processing constants stay in Table S7, and thresholds left to the released code are not listed.

**Table S9.** Constants of the production run, with the section that uses each.

| Constant | Value | Section |
|---|---|---|
| Page image | 150 dpi, shrunk if larger to 1,600 pixels on the longer side | S2.1 |
| Panel padding | 12% of the returned box on every side | S2.1 |
| Panel image | About 1,500 pixels wide, at 220 to 900 dpi | S2.1 |
| Dark pixel of the plot frame | Gray level below 128 of 255 | S2.3 |
| Outlier pairs | At most two dropped, each only if its residual exceeds both 0.3° and three times the median residual | S2.4 |
| Placement of the LLM's values | RMSE below 0.25° | S2.4 |
| Placement rule (ii) | Within 0.5° of every OCR label, if the next best misses some label by more than 1.0° | S2.4 |
| Placement rule (iii) | Labels span at least 75% of the marks | S2.4 |
| Fit checks, range | Supporting points widened by 10% on each side | S2.4 |
| Fit checks, axis | Rises from left to right, stays within −5° to 130° and spans 1.5° to 135° | S2.4 |
| Fit checks, RMSE | Below 0.5° for the OCR fit and 0.35° for the LLM fit | S2.4 |
| Consensus | Largest 2θ difference at most 0.5° at 24 evenly spaced columns across the plot frame | S2.4 |
| Lone-fit rule | At least three ticks and an RMSE of at most 0.15° | S2.4 |
| Axis arbitration | Single placement on the tick marks, plausible axis and RMSE below 0.35° | S2.4 |
| Anchor | At least 8 points | S2.5 |
| Ink, dark pixel | Gray level $g$ below $\max(80,\ g_{75} - 45)$, with $g_{75}$ the frame's 75th-percentile gray level | S2.5 |
| Ink, colored pixel | Saturation above 60 and $g < 235$ | S2.5 |
| Search half-width, color and ink masks | $w = \max(4, \lfloor 0.045\,H \rfloor)$ pixels | S2.5 |
| Search half-width, wide mask | $w = \max(4, \lfloor 0.09\,H \rfloor)$ pixels | S2.5 |
| Running median | Trace heights more than $\max(6,\ 0.05\,H)$ pixels from a 5-column running median replaced by it | S2.5 |
| Tolerance | $\tau = \max(3, \lfloor 0.02\,H \rfloor)$ pixels | S2.5 |
| Band half-width | $b = \max(6, \lfloor 0.10\,H \rfloor)$ pixels | S2.5 |
| Acceptance gates | $s \geq 0.55$, $p \geq 0.70$, $r \geq 0.60$ and $c \geq 0.55$ | S2.5 |

# Section S3. Prompts and Output Schemas

## S3.1. Request Format

Each request of the four LLM calls carries a system prompt, a user prompt, one PNG image and a strict JSON output schema (Table S10; Figure S3). In the run, a request timed out after 240 s and was tried up to four times on transient errors.

**Table S10.** The four LLM calls of the production run.

| Call | LLM tasks carried | When it runs | Image sent | Text sent | Calls[a] |
|---|---|---|---|---|---|
| Panel location | Panel location | Once per PDF page of the 3,954 processed articles | Rendered page (Section S2.1) | Page text, first 2,500 characters | ≈ 37,489 |
| Figure analysis | Diffraction-pattern check, curve enumeration, tick reading | Once per figure attempt, and again after a reframe (Section S2.2) | Panel re-rendered about 1,500 px wide (Section S2.1) | Page text, first 1,500 characters | ≥ 5,292 |
| Anchor tracing | Anchor tracing | Once per figure attempt with an accepted calibration | Panel image with a red 10 × 10 grid on the plot frame | List of measured curves | ≥ 4,730 |
| Axis arbitratio n | Axis arbitration | Only when calibration accepts no fit (Section S2.4) | Strip along the x axis, magnified three times | Tick values read by the LLM and by OCR; number of CV tick marks | ≥ 558 |
| Total | | | | | ≥ 48,069 |

[a] Panel location made one call per PDF page. It covered 24,086 pages of the 2,554 articles in the run and 13,403 pages of the 1,400 articles without a located panel (Table S2). A few final pages of the 6 articles in progress at the stop may never have been sent. The other counts are lower bounds, because repeats on a re-rendered panel went unrecorded. Figure analysis answered 5,292 of the 5,295 figure attempts, and the other 3 got no answer because of provider quota errors.

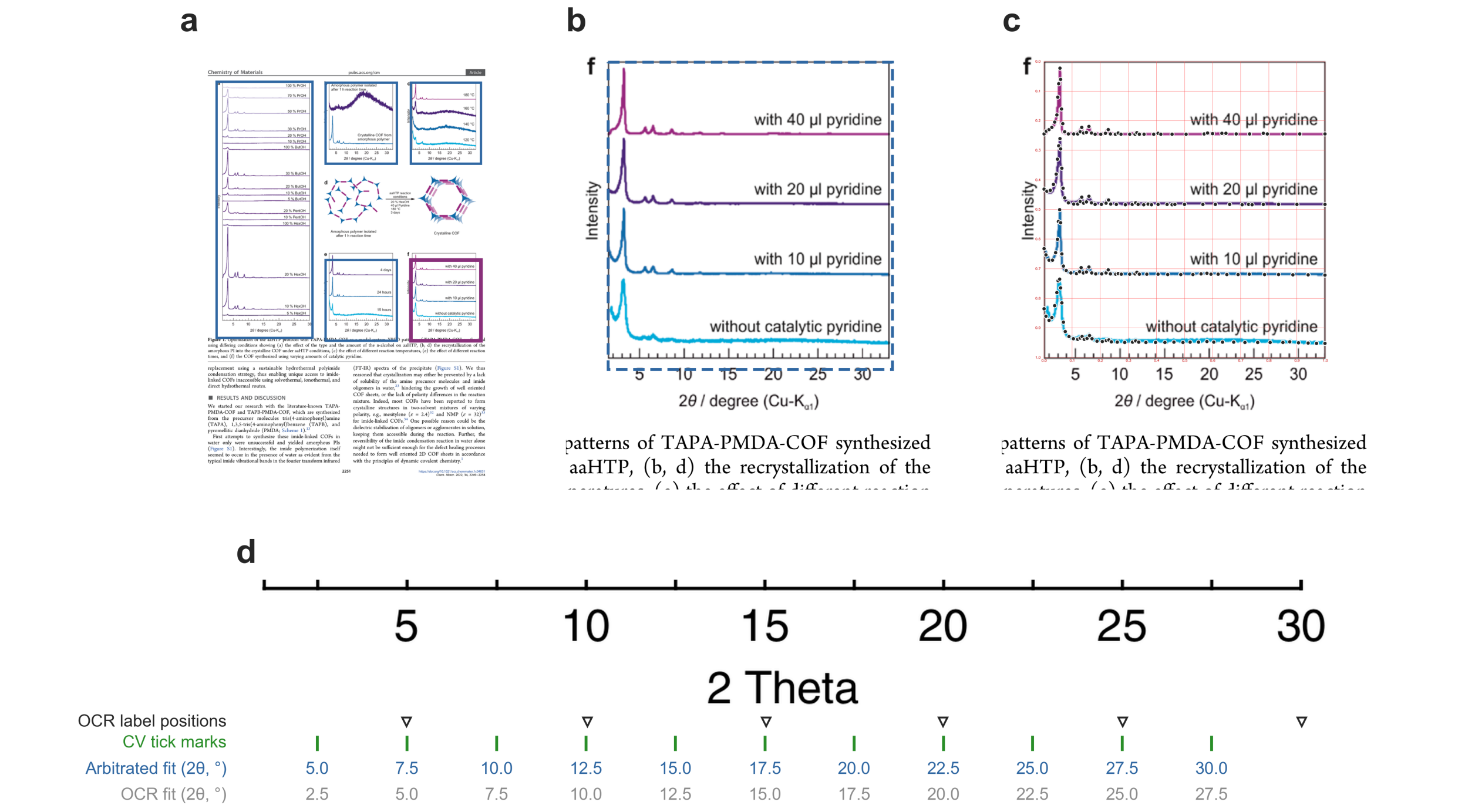

**Figure S3. Images sent to the four LLM calls, with their answers.** (a) Panel location, with returned boxes; (b) figure analysis of panel f, with the LLM's plot box (dashed); (c) anchor tracing, with grid and returned anchors; and (d) axis arbitration, with the strip as sent, whose arbitrated axis is one tick step (2.5°) too high. Panels a–c are adapted from page 2251 and Figure 1f of ref. [9] (© 2022 The Authors) and panel d from Figure 4e of ref. [10] (© The Author(s) 2018), both under a CC BY 4.0 license, with the boxes, grid, anchors, and tick annotations added and the axis strip magnified.

## S3.2. Panel Location

**Box S1.** Panel-location prompts and output schema, where {PAGE_TEXT} holds the first 2,500 characters of the page text (empty if none). Each returned box becomes one figure attempt, with panel_label stored as the figure label. The fields confidence and notes are not used. In all four boxes, bracketed headings were not sent, and a wrapped line is one prompt line.

---

*[System prompt]*
*You locate PXRD diffraction plot panels on scientific paper pages. Return strict JSON. Fractional coordi nates, full axis labels included in each box. If the page has no PXRD plot, return an empty list.*

*[User prompt]*
*Find every powder X-ray diffraction (PXRD/XRD) plot panel on this page.*
*A PXRD panel is a 2D line plot whose x-axis is 2-theta (2θ, degrees, Bragg angle). Include each individu al panel of multi-panel figures separately. EXCLUDE: SEM/TEM images, photographs, adsorption isothe rms, IR/Raman/NMR/UV-Vis spectra, TGA curves, schematics, and structures.*

*For each panel return its bounding box in FRACTIONS of the page (x0,y0 top-left, x1,y1 bottom-right, 0.. 1). The box must include the full plot frame plus its axis tick labels and axis titles, but as little surroundin g content as possible. Also return the panel letter label if printed (a, b, c...).*

*Page text excerpt (may help identify figures):*
*{PAGE_TEXT}*

*[Output schema]*
```
class PanelBox(BaseModel):
    """One PXRD panel located on a page, in fractional page coordinates."""

    x0: float = Field(ge=0.0, le=1.0)
    y0: float = Field(ge=0.0, le=1.0)
    x1: float = Field(ge=0.0, le=1.0)
    y1: float = Field(ge=0.0, le=1.0)
    panel_label: Optional[str] = None
    confidence: float = Field(default=0.0, ge=0.0, le=1.0)

class PagePanels(BaseModel):
    """All PXRD panels the model can see on one rendered page."""

    panels: List[PanelBox] = Field(default_factory=list)
    notes: str = ""
```

---

## S3.3. Figure Analysis

**Box S2.** Figure-analysis prompts and output schema, where {PAGE_TEXT} holds the first 1,500 characters of the page text, or N/A. Every request sent N/A as the caption. The pipeline uses is_pxrd, reason, plot_box and x_tick_values, and for each curve label, material_name, sample_state, color and is_stick_pattern (Sections S2.2 to S2.5). It stores the other fields without using them.

---

*[System prompt]*
*You analyze PXRD figure crops. Return strict JSON. Read tick labels and curve labels EXACTLY as print ed; never invent values that are not visible. Curves ordered top to bottom.*

*[User prompt]*
*Analyze this figure crop from a chemistry paper.*

*1. is_pxrd: is it a powder XRD diffractogram (x-axis = 2θ in degrees)? Plots of adsorption, spectroscopy, TGA etc. are NOT PXRD.*
*1b. plot_box: the bounding box of the PXRD plot's AXES FRAME only (the rectangle delimited by the axi s lines, excluding tick labels, axis titles, and any neighboring drawings/panels), in fractions of this image (x0,y0 top-left to x1,y1 bottom-right).*
*2. curves: list every REAL diffraction trace, ordered top to bottom at their left end. For each: the printed l abel exactly as written (legend or inline text), material name, sample state (experimental/simulated/calcin ed/...), and its line color as a simple English word. Mark stick/bar reference patterns with is_stick_patter n=true. Do NOT count Bragg-position tick rows or axis lines as curves.*
*3. x_tick_values: ALL numeric tick labels printed along the x-axis, left to right, exactly as printed (e.g. [5, 10,15,20,25,30]).*
*4. x_axis_title / y_axis_title exactly as printed; x_unit_is_degrees.*
*5. grayscale_only: true when all curves are black/gray.*
*6. has_inset: true if a smaller plot is embedded inside.*

*Caption: N/A*
*Context: {PAGE_TEXT}*

*[Output schema]*
```
class CurveSpec(BaseModel):
    """Semantic description of one diffraction trace inside a panel."""

    order_from_top: int = Field(default=1, ge=1)
    label: Optional[str] = None
    material_name: Optional[str] = None
    sample_state: Optional[str] = None
    color: str = "black"
    is_stick_pattern: bool = False
    confidence: float = Field(default=0.0, ge=0.0, le=1.0)

class FracBox(BaseModel):
    """Axis-aligned box in fractional image coordinates."""

    x0: float = Field(ge=0.0, le=1.0)
```

```
    y0: float = Field(ge=0.0, le=1.0)
    x1: float = Field(ge=0.0, le=1.0)
    y1: float = Field(ge=0.0, le=1.0)

class FigureAnalysis(BaseModel):
    """One-shot deep reading of a panel crop: classification + inventory + axis."""

    is_pxrd: bool
    confidence: float = Field(default=0.0, ge=0.0, le=1.0)
    reason: str = ""
    figure_label: Optional[str] = None
    plot_box: Optional[FracBox] = None
    curves: List[CurveSpec] = Field(default_factory=list)
    x_tick_values: List[float] = Field(default_factory=list)
    x_axis_title: str = ""
    x_unit_is_degrees: bool = True
    y_axis_title: str = ""
    grayscale_only: bool = False
    has_inset: bool = False
    caption_text: str = ""
```

---

## S3.4. Anchor Tracing

**Box S3.** Anchor-tracing prompts and output schema, with the grid drawn on the CV plot frame. The prompt calls this frame the plot box. {N} counts the measured curves, and {CURVE_LIST} gives each curve on one line with its number, color, label (or material name) and sample state, as for Figure S3c. The pipeline uses points, together with label where figure analysis gave none (1,478 accepted curves).

---

```
[System prompt]
You digitize curve shapes from plots into fractional-coordinate polylines. Return strict JSON. y_frac is m
easured DOWNWARD from the top of the plot box. Peaks must not be skipped.

[User prompt]
This PXRD figure crop has a red 10x10 reference grid drawn over the plot box; the grid spans the plot int
erior with fractional labels 0.0 to 1.0 on both axes (x: left→right, y: TOP→bottom, so y=0.0 is the top ed
ge of the plot box and y=1.0 the bottom edge).

Trace ALL {N} diffraction curve(s), ordered top to bottom:
{CURVE_LIST}

For each curve return a polyline of 40-80 points (x_frac, y_frac) following the visible trace from its left e
nd to its right end. CRITICAL accuracy requirements:
- Use the red gridlines to estimate coordinates precisely.
- Capture every visible peak: at each sharp peak place a point at the apex plus points on both flanks.
- Between peaks follow the baseline level of THAT curve (stacked curves have different baseline heights
— do not drift to a neighbor).
- Densify points around peaks; sparse points are fine on flat regions.

[Output schema]
class AnchorPoint(BaseModel):
    x_frac: float = Field(ge=-0.05, le=1.05)
    y_frac: float = Field(ge=-0.05, le=1.05)

class CurveAnchor(BaseModel):
    """Coarse polyline for one curve in fractional plot-box coordinates.

    (0,0) is the top-left of the plot interior, (1,1) the bottom-right.
    """

    order_from_top: int = Field(default=1, ge=1)
    label: Optional[str] = None
    points: List[AnchorPoint] = Field(default_factory=list)
    confidence: float = Field(default=0.0, ge=0.0, le=1.0)

class CurveAnchorSet(BaseModel):
    curves: List[CurveAnchor] = Field(default_factory=list)
    notes: str = ""
```

---

## S3.5. Axis Arbitration

**Box S4.** Axis-arbitration prompts and output schema, where {CANDIDATES} is one line. For Figure S3d, this line read candidate tick value lists: LLM=[5.0, 10.0, 15.0, 20.0, 25.0, 30.0], OCR=[5.0, 10.0, 15.0, 20.0, 25.0, 30.0]; CV found 11 tick marks. The pipeline uses tick_values, or the first and last values when that list is empty, and ignores confidence and notes.

---

*[System prompt]*
*You resolve conflicting axis readings by carefully reading magnified tick labels. Return strict JSON.*

*[User prompt]*
*This is a magnified x-axis strip of a PXRD plot. Our automatic calibration produced conflicting readings: {CANDIDATES}*

*Read the axis carefully and report: the numeric value of the FIRST (leftmost) tick label, the LAST (rightm ost) tick label, and the full ordered list of tick label values. Only report what is printed.*

*[Output schema]*

```
class CalibrationArbitration(BaseModel):
    """LLM arbitration when calibration candidates disagree."""

    first_tick_value: Optional[float] = None
    last_tick_value: Optional[float] = None
    tick_values: List[float] = Field(default_factory=list)
    confidence: float = Field(default=0.0, ge=0.0, le=1.0)
    notes: str = ""
```

---

**Same tick values.** Of the 165 accepted arbitrations, 157 followed two fits that disagreed. In 150 of these, the LLM reread the same tick values. The new line then repeats the earlier LLM fit to within 0.05°.

**Contradicted lines.** In 124 of the 150, the printed labels contradict the line. A contradiction means that at least two labels lie within a quarter tick spacing of a CV tick mark. The line also puts each such mark more than 0.5° from its label. The labels confirm the line in 24, and 2 are mixed.

**Offset size.** In 123 of the 124, the line sits exactly one CV tick spacing from the labels. This offset is 2.5° in 86, about 5° in 30, and 10° in 3. In the other 4, the offset is another tick spacing. In all but one of the 124, the line is too high.

**Check by eye.** A random sample of 12 of the 124 figures, stratified by publisher, was checked by eye. All 12 are one tick step off.

# Section S4. Per-Task Ablation

This section compares four LLM tasks with classical baselines, one task at a time. Section S4.1 gives the design and the measures, and Sections S4.2 to S4.5 report the four tasks in turn.

## S4.1. Design and Measures

**Design.** Each ablation test replaces the LLM's output for one task with a baseline's and keeps all other LLM outputs (Table S11). The kept outputs make the baselines stronger than a purely classical system.

**Table S11.** The four ablated tasks and the LLM outputs that each baseline keeps.

| LLM task replaced | Baselines | Still taken from the LLM | Scored on |
|---|---|---|---|
| Panel location | Ink projection, embedded image, vector drawing | Starting region of ink projection: the image or vector group that holds the LLM's box | 4,477 accepted figures (2,317 papers) |
| Curve enumeration | Hue clustering | Plot box; grayscale flag | 4,654 stored figures[a] |
| Tick reading | OCR, in three calibration variants | Traced curve and its lowest stored peak | 1,186 curves (445 papers)[b] |
| Anchor tracing | Eight anchor baselines | Curve list; curve color (in the anchor for four baselines, in ink snapping for seven); number of traced curves and rank (two baselines) | 1,175 curves (444 papers)[b] |

[a] The other 10 of the 4,664 stored figures lack a plot box (2) or have a plot box under 40 × 30 pixels or an unreadable crop (8).

[b] For tick reading, accepted curves with at least one stored peak and a high-confidence link to a record with a parsable peak list; for anchor tracing, the last row of Table S4 (Section S1.3).

**Reference box.** For panel location, the reference box $T$ is the stored panel box without the pipeline's 12% margin. For the 914 accepted figures whose $T$ came from a reframe (Section S2.2), rescoring against the LLM's panel-location box changes no verdict or ranking.

**Outcome classes.** The intersection over union (IoU) measures how well two boxes agree, as their overlap area divided by their combined area. A box $B$ covers $T$ when $|B \cap T| \geq 0.9\,|T|$. Each figure takes the first of four classes that applies. It is a tight box if the best IoU with $T$ is at least 0.5 [11]. Otherwise, it is an oversized box if the smallest covering box has at least twice the area of $T$, and a loose box if that box is smaller. It is missed if no box covers $T$.

**First-peak difference.** Tick reading and anchor tracing score each curve $c$ with a high-confidence link by its first-peak difference. This difference is $e_c = |\theta_c - \theta_c^{\text{text}}|$, with $\theta_c^{\text{text}}$ the first reported peak (Section S1.3).

**Two differences.** Two differences compare the LLM ($L$) with a baseline ($B$). The first, $\Delta_{\text{paired}}$, asks which method lands closer to the text on the same curves in a typical paper, with each paper counting once. The second, $\Delta_{\text{med}}$, asks which method has the lower typical error over all shared curves. Negative values of both favor the LLM. With $C$ the curves both cover and $C_p$ those from paper $p$, they are

$$\Delta_{\text{paired}} = \underset{p}{\text{median}}\left[\underset{c\in C_p}{\text{median}}(e_c^L - e_c^B)\right], \qquad \Delta_{\text{med}} = \underset{c\in C}{\text{median}} e_c^L - \underset{c\in C}{\text{median}} e_c^B.$$

## S4.2. Panel Location

**Baselines.** The three panel-location baselines each read the article PDF and may return many boxes, of which scoring keeps the best match to $T$ (Table S12). The embedded image returns the placement rectangle of each raster image on the page. The vector drawing returns the bounding boxes of the page's vector drawing elements, merged when close.

**Ink projection.** Ink projection [12], [13], [14] splits a region of the page at blank gaps. It starts from the smallest embedded image or vector group that covers $T$. It cuts this region once along columns, then once along rows, wherever a blank run reaches the gap threshold.

**Table S12.** Outcomes of the panel-location baselines on the 4,477 accepted figures (2,317 papers). The gap threshold is the narrowest blank run at which ink projection cuts, as a share of the region's width (columns) or height (rows), and at least 3 pixels. "Region not cut" counts starting regions returned as one piece.

| Baseline[a] | Region not cut | Tight box | Oversized box | Loose box | Missed |
|---|---|---|---|---|---|
| *As used (gap threshold 3.5%)* | | | | | |
| Ink projection | 3,653 (81.6%) | 472 (10.5%) | 3,765 (84.1%) | 87 (1.9%) | 153 (3.4%) |
| Embedded image | – | 368 (8.2%) | 3,256 (72.7%) | 79 (1.8%) | 774 (17.3%) |
| Vector drawing | – | 135 (3.0%) | 174 (3.9%) | 5 (0.1%) | 4,163 (93.0%) |
| *Ink projection, narrower gap threshold* | | | | | |
| 0.5% | 395 (8.8%) | 1,957 (43.7%) | 1,331 (29.7%) | 56 (1.3%) | 1,133 (25.3%) |

[a] Vector elements (at most the 4,000 largest per page) merge within 8 points, and embedded images and merged groups under 400 square points are ignored. Ink projection renders its region at 144 dpi, counts pixels darker than 200 (of 255) as ink, drops pieces under 5% of the region, and starts from the whole page when no image or group covers $T$.

**Embedded-image failures.** The embedded image fails because publishers embed a composite figure as one picture [13]. A single-letter panel label appears in 4,058 of the 4,477 figures (90.6%). The oversized boxes of this baseline had a median of 6.4 times the panel's area (Figure S4a).

**Other failures.** The vector drawing fails because most plots are raster images. Of the figures, 4,035 (90.1%) lie on a page with fewer than 50 drawing elements. Ink projection fails because the gap between two panels is often no wider than gaps inside a panel. Its uncut regions gave 3,336 of its 3,765 oversized boxes.

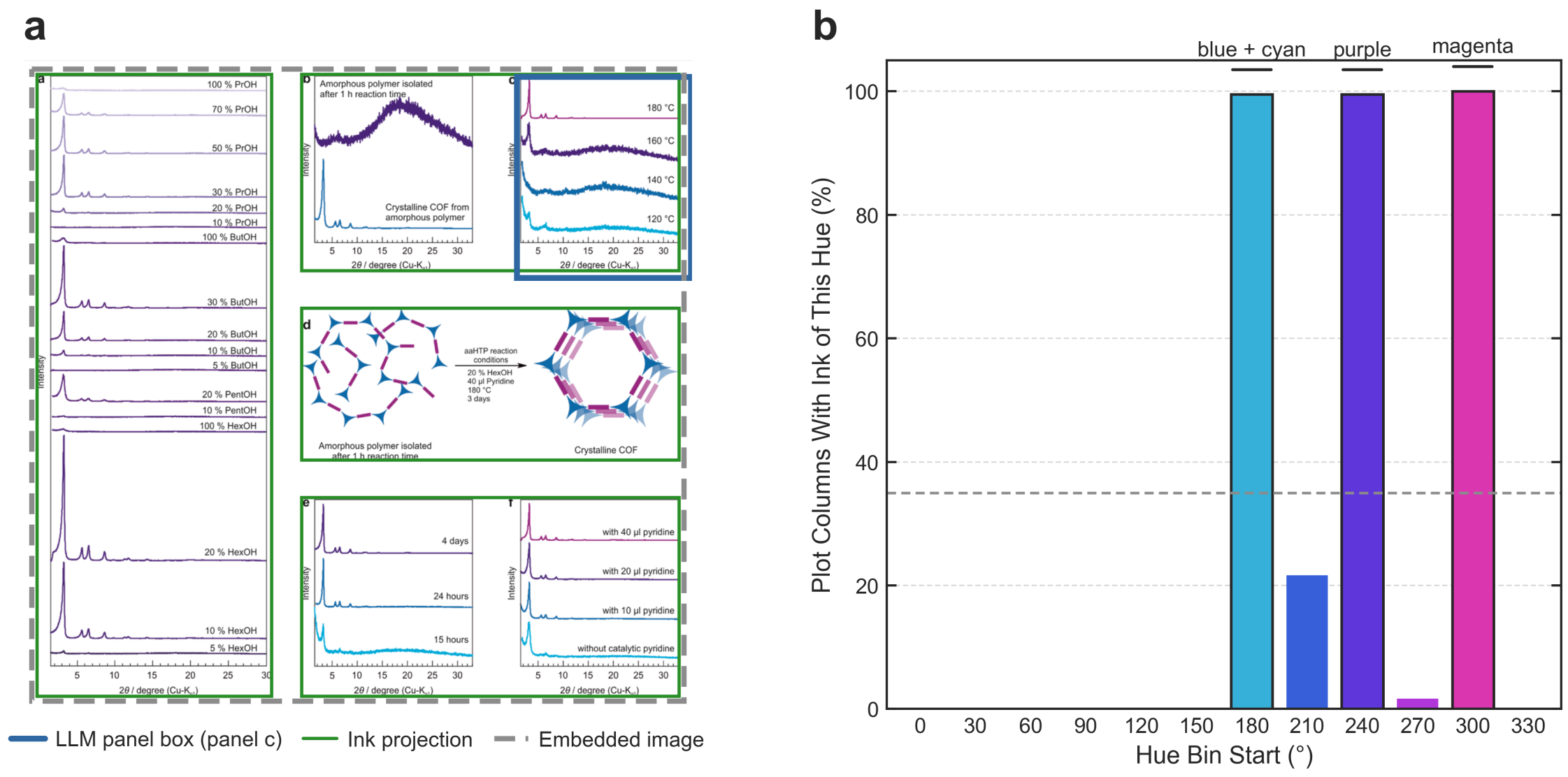

**Figure S4. Classical baselines applied to one composite figure.** (a) Panel location, with the LLM's box (blue), embedded image (gray dashed; IoU 0.08) and ink-projection pieces (green; best IoU 0.41); (b) curve enumeration, with the columns holding ink per 30° hue bin, kept bins outlined. Blue and cyan share a bin, so three curves are counted, not four. Panel a is adapted from Figure 1 of ref. [9] (© 2022 The Authors) under a CC BY 4.0 license, with the boxes added.

## S4.3. Curve Enumeration

**Hue clustering.** The hue-clustering baseline counts the curves inside the LLM's plot box in six steps (Table S13). It describes each pixel by its hue, saturation and value (HSV), which give the pixel's position on the hue wheel, its vividness and its brightness.

**(i) Crop.** The baseline crops the panel to the plot box, shrunk to at most 600 pixels wide.

**(ii) Hue bins.** It assigns each colored ink pixel to one of 12 hue bins of 30°. A pixel counts as colored ink when its saturation is at least 0.25 and its value at least 0.12.

**(iii) Dark ink.** On grayscale panels only, it adds one cluster of dark ink, with lower saturation and a value of 0.02 to 0.55.

**(iv) Keep.** It keeps a bin or cluster with at least 0.08% of the plot's pixels and ink in at least 35% of its columns.

**(v) Merge.** It merges kept bins that neighbor each other on the hue wheel.

**(vi) Count.** It counts the merged groups and returns zero for a plot with fewer than 50 ink pixels.

**Table S13.** Curve counts of the hue-clustering baseline on 4,654 stored figures. "Same", "Fewer" and "More" compare the baseline's count with the number of curves the LLM lists, stick patterns excluded. The variant adds the dark-ink cluster of step (iii) on color panels.

| Panels and baseline | Panels | Same | Fewer | More |
|---|---|---|---|---|
| Color, all | 4,382 | 22.7% | 73.9% | 3.4% |
| Color, LLM lists a black or gray curve | 2,877 | 9.8% | 89.2% | 1.0% |
| Color, LLM lists no black or gray curve | 1,505 | 47.3% | 44.8% | 7.9% |
| Color, variant that also counts dark ink | 4,382 | 40.1% | 35.5% | 24.4% |
| Grayscale | 272 | 51.5% | 41.2% | 7.4% |

**Undercounts.** Dark ink fully explains 1,513 (46.7%) of the 3,239 undercounts on color panels. The baseline ignores dark ink there (step iii), and on these panels it counts at least the LLM's non-dark curves. The other 1,726 undercounts involve colored curves that merged in one hue bin or in neighboring bins, or were lost (Figure S4b). On grayscale panels, all 140 matches are on the 162 panels where the LLM lists one curve.

## S4.4. Tick Reading

**Variants.** Each of the four calibration variants converts the pixel column of a curve's lowest stored peak with its own calibration line (Table S14). The variants differ only in the source of the tick values, the LLM or OCR, and in the source of their positions. The OCR-only fit is the OCR fit of Section S2.4, rebuilt without its RMSE caps.

**Table S14.** Calibration variants on the 1,186-curve cohort (445 papers), and paired comparisons between them. Medians and $\Delta_{\mathrm{paired}}$ are first-peak differences in degrees; negative $\Delta_{\mathrm{paired}}$ favors A. "> 5°" is the share of covered curves beyond 5° from the text.

| Variant, or comparison A − B | Curves (papers) | Coverage[a] | Median or $\Delta_{\mathrm{paired}}$ (95% CI) | > 5° |
|---|---|---|---|---|
| *Variants: median first-peak difference* | | | | |
| Pipeline calibration (fit accepted in production) | 1,186 (445) | 100.0% | 0.083 (0.071 to 0.097) | 9.7% |
| LLM tick values on CV tick marks | 550 (235) | 46.4% | 0.082 (0.065 to 0.097) | 8.9% |
| OCR tick values on CV tick marks | 510 (228) | 43.0% | 0.083 (0.067 to 0.098) | 9.4% |
| OCR-only fit: OCR values at OCR label positions | 1,159 (439) | 97.7% | 0.113 (0.102 to 0.127) | 10.2% |
| *Paired comparisons:* $\Delta_{\mathrm{paired}}$ | | | | |
| Pipeline calibration − OCR-only fit | 1,159 (439) | – | 0.000 (−0.009 to 0.000) | – |
| LLM tick values − OCR-only fit | 526 (229) | – | −0.016 (−0.029 to −0.005) | – |
| OCR tick values − OCR-only fit | 510 (228) | – | −0.014 (−0.028 to −0.003) | – |
| LLM tick values − OCR tick values | 493 (218) | – | 0.000 (0.000 to 0.000) | – |

[a] Share of the cohort with a usable fit. A usable fit passes the fit checks of Section S2.4 without the RMSE caps, over the plot's extent rebuilt from the traced curves' 2θ range. Values were paired with CV tick marks by the rules of Section S2.4, without the tie-breaks that use OCR.

**Tick sets.** The tick sets of the LLM and of OCR are identical on 4,196 of the 4,609 stored figures on which both read at least two values. Two tick sets count as identical when they hold the same distinct values.

**Ties.** The pipeline calibration and the OCR-only fit tie exactly in 84 of the 439 papers (main text Section 3.2). All 84 hold curves on which the pipeline calibration is the OCR fit itself. Over all papers, this holds for 261 of the 1,159 curves. On 119 of them the OCR fit was the lone fit, and on 142 it was the consensus fit with the smaller RMSE (Section S2.4).

## S4.5. Anchor Tracing

**Baselines.** The eight anchor baselines each replace only the LLM's coarse anchor (Table S15). Each places one height per pixel column, and these heights are smoothed and sampled into an anchor. The released code then does the precise work, from ink snapping and the gates (Section S2.5) to first-peak detection.

**Matched processing.** The run's records kept each final trace but not the LLM's anchors, which remain only in the response cache. The LLM side therefore uses the "LLM replay", its production trace read back from the overlay image. The replay enters at the calibration step without new gates. Both sides thus share the plot frame rebuilt from the overlay image, the calibration, the clipping and the peak finding, which we term "matched processing".

**Negative control.** Coverage shows only that some trace passed the gates. In a negative control, an anchor taken from an earlier figure still passed on 51.5% of 689 figures. These were all the figures of the earlier 1,206-curve cohort except the first, which had no earlier figure. That cohort used the selection rule of the 1,175-curve cohort, applied while the curves of Section S3.5 still counted as clean (Section S7.2).

**Table S15.** The eight anchor baselines under matched processing, against the LLM replay (1,175 curves, 444 papers; replay coverage 98.5%). Coverage differences are in percentage points (replay minus baseline), and $\Delta_{\text{paired}}$ and $\Delta_{\text{med}}$ in degrees on the shared curves. Positive coverage and negative accuracy differences favor the LLM.

| Baseline | Anchor height in each pixel column[a] | Taken from the LLM[b] | Shared curves (papers) | Coverage difference (95% CI) | $\Delta_{\text{paired}}$ (95% CI) | $\Delta_{\text{med}}$ (95% CI) |
|---|---|---|---|---|---|---|
| Color centroid[d] | Ink-weighted mean row of ink in the curve's color | Color | 995 (397) | 12.8 (10.4 to 15.2) | −0.020 (−0.035 to −0.001) | +0.038 (−0.071 to +0.234) |
| Centroid, unfiltered | As color centroid, without the running median | Color | 986 (396) | 13.5 (11.1 to 16.1) | −0.019 (−0.032 to −0.002) | +0.048 (−0.063 to +0.258) |
| Color top edge | Topmost ink pixel in the curve's color | Color | 934 (394) | 18.1 (15.5 to 20.9) | −0.025 (−0.049 to −0.007) | −0.002 (−0.114 to +0.154) |
| Top edge, unfiltered | As color top edge, without the running median | Color | 938 (394) | 17.8 (15.2 to 20.4) | −0.023 (−0.054 to −0.008) | −0.028 (−0.134 to +0.131) |
| Color clustering[c] | Mean row of color cluster $k$ of all ink (k-means, $\min(n, 8)$ clusters, ranked from the top) | $n$, $k$; color in snapping | 802 (337) | 29.3 (25.8 to 32.9) | −0.109 (−0.160 to −0.065) | −0.152 (−0.285 to +0.039) |

| Baseline | Anchor height in each pixel column[a] | Taken from the LLM[b] | Shared curves (papers) | Coverage difference (95% CI) | $\Delta_{\text{paired}}$ (95% CI) | $\Delta_{\text{med}}$ (95% CI) |
|---|---|---|---|---|---|---|
| Ink centroid, strict | Ink-weighted mean row of all ink | Nothing | 780 (309) | 31.2 (26.8 to 35.7) | −0.100 (−0.140 to −0.060) | −0.132 (−0.491 to +0.088) |
| Ink centroid | Ink-weighted mean row of all ink | Color in snapping | 763 (311) | 32.7 (28.3 to 37.1) | −0.100 (−0.139 to −0.057) | −0.137 (−0.478 to +0.079) |
| Horizontal band | Ink-weighted mean row of all ink in band $k$ of $n$ equal bands | $n$, $k$; color in snapping | 759 (344) | 32.9 (29.7 to 36.1) | −0.142 (−0.211 to −0.100) | −0.375 (−0.601 to −0.170) |

[a] Ink weighting weights each pixel by its darkness, $255 - g$ for gray level $g$. Except in the unfiltered baselines, the heights are smoothed by a running median over $\max(5, \lfloor w/100 \rfloor)$ columns, made odd, for a plot $w$ pixels wide. Every anchor has 60 evenly spaced points, and fewer than 8 inked columns give no anchor.

[b] Color is the curve color from curve enumeration, used in the anchor and in ink snapping. An unrecognized color, or one without ink, fell back to all ink for 77 of the 1,318 traced curves in the cohort's 673 figures. Here $n$ is the number of traced curves in a figure and $k$ the curve's rank from the top. The baselines that take $n$ and $k$ reduce to the ink centroid when $n = 1$, as in 390 of these figures.

[c] The k-means step [15] starts from unseeded centers; a rerun gave 68.1% coverage instead of 69.2%.

[d] For the color centroid, 223 of 397 papers favor the LLM, but the median errors over the shared curves are 0.330° (replay) and 0.292° (baseline).

**Coverage bias.** The cohort favors the LLM, because it holds only curves that the LLM traced in production. An oracle anchor, drawn from the color and vertical range of the LLM's final trace, passed the gates on 94.0% of curves, 53 curves fewer than the replay. Six variants built later, which are not in Table S15, passed on at most 74.8%. A stricter count treats the 45 curves lost in production as LLM failures and baseline successes, and it removes those 53 curves from the replay. The LLM then still covers 90.5% of curves, against 86.2% for the color centroid.

**Original plan.** The original written plan (Section S1.4) compared the LLM's stored first peaks with each baseline under matched processing. It favored the LLM against all eight baselines. Yet the stored LLM curves also beat the LLM replay, the same traces processed twice, by 0.249° (95% CI, 0.133° to 0.491°). This comparison used the 1,188 curves (450 papers) of the earlier cohort with a first peak from both the stored curve and the replay. The plan's advantage thus came from the processing, not from the anchor.

# Section S5. Accuracy of the Extracted Curves

This section compares the accepted curves with the authors' text (Section S5.1), with a hand digitization (Section S5.2) and with synthetic panels of known truth (Section S5.3).

## S5.1. Comparison with the Authors' Text

**Curves compared.** The text comparison scores the first peak of the 1,175 curves in the last row of Table S4 (Section S1.3). The parsed text value and its link can both be wrong (Section S1.3), so the agreement is not an error estimate.

**Negative control.** The negative control of main-text Table 3 pools all 1,375,010 cross-paper curve–value pairs and has a median difference of 1.52°. This summary takes each curve's median first, which allows a paper-level CI. The control median is then 1.17° (95% CI, 1.05° to 1.26°), and the mean share within 0.3° is 13.0% (95% CI, 12.1% to 13.9%).

**All reported angles.** A second comparison matches every reported angle to the nearest digitized peak of the same material (Figure S5). It counts only angles within the material's mean plotted range ±0.5°. A peak's prominence is how far it rises above the higher of its two flanking valleys. Digitized peaks, found with SciPy, need a prominence of at least 3% of the intensity range. The control uses the peaks and plotted range of a random partner from another paper.

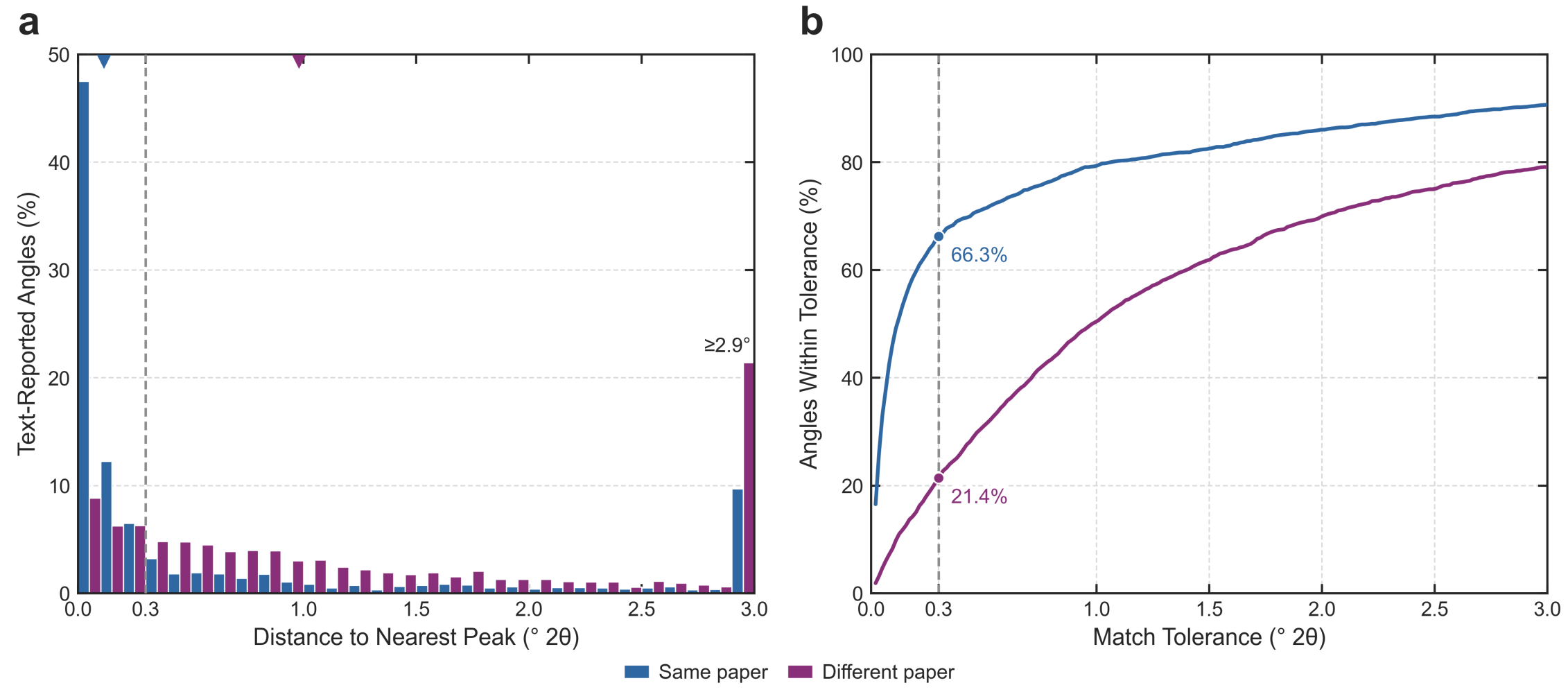

**Figure S5. Digitized peaks against the angles reported in the text for 668 materials from 447 papers.** (a) Absolute difference between each reported 2θ angle and the nearest digitized peak, of the same material (blue; 2,401 angles) or of the control partner (purple; 2,282 angles); triangles, medians (0.115° and 0.980°); dashed, the 0.3° match tolerance; last bin, 2.9° or more. (b) Share of reported angles matched against the match tolerance; at 0.3°, 66.3% (95% CI, 63.6% to 68.9%) of same-paper angles match, against 21.4% (95% CI, 19.4% to 23.5%) for the control.

**Large differences.** Of the 1,175 curves, 102 (8.7%) place their first peak more than 5° from the first reported peak (Table S16). On the 1,186 curves of Section S4.4, which take the lowest stored peak as first, this tail is 9.7%.

**Causes.** In 26 curves, the detector took as first a peak at a higher angle than the first reported one (15 curves) or an unlisted peak at a lower angle (11). Of the 38 curves with no match, 16 have a reported peak list that starts at the stacking peak (26.1–28.3°) [16]. Such a list likely omits low-angle reflections. Of the 33 curves whose first reported peak lies below the plotted range, 8 curves from 5 figures are inset curves converted with the panel's main axis. Of the rest, 3 are inspected main-panel curves, and 22 were not inspected.

**Table S16.** Curves of the text comparison whose first peak lies more than 5° from the first reported peak, each counted in the first row that applies. A match lies within 0.3° of a digitized peak of the same curve; Tier B counts links made by elimination (Section S1.3).

| What the curve and the text show | Curves | Tier B |
|---|---|---|
| First reported peak lies below the start of the curve's plotted range | 33 | 9 |
| First reported peak matches a digitized peak, but the detector took another peak as first | 26 | 4 |
| Another reported angle matches a digitized peak | 5 | 0 |
| No reported angle matches any digitized peak | 38 | 19 |
| Total | 102 | 32 |

## S5.2. Hand Digitization

**Sampling.** The hand digitization uses 60 panels from 55 papers, drawn from the stored figures with a seed fixed by the written plan (Section S1.4). They fall in six strata that cross color or grayscale with 1, 2–3 or at least 4 accepted curves. Grayscale panels, 5.8% of the stored figures, fill 12 of the 60 panels, so medians are weighted by each stratum's share.

**Digitization.** One author (S.G.) acted as the operator and digitized every curve in WebPlotDigitizer [17]. Its automatic color extraction traced each curve, and the operator placed points by hand where the extracted trace left the curve. The first pass took 2.61 h, excluding the time spent choosing curves. Because the color extraction and ink snapping both follow the drawn ink of one color, the two can share errors, for example where two curves of the same color cross.

**Blinding and selection.** Panel crops came in random order under anonymous codes, without pipeline traces or paper identities. Calculated, refined, reference and stick patterns were skipped, as were curves inside insets. On 38 panels, the operator was told which curves to skip, as read from the in-panel labels or, on 6, from the caption. This reading was not blinded to the pipeline output, so the curve count on these panels is not blinded.

**Curve counts.** On the other 22, the operator chose the curves and, on 3, also had the caption. The curve counts agreed on all 22, and all 7 count disagreements (Table S17) lie among the 38. Panel V055 also came with its shared 2θ axis, which the crop had lost.

**Table S17.** Panels on which the operator's and the pipeline's counts of measured curves differ, with the cause of each; the pipeline count is of accepted curves.

| Panel | Operator | Pipeline | Cause |
|---|---|---|---|
| V005 | 1 | 3 | Calculated curves kept: two "-Sim" curves escaped the run's keyword filter (Section S2.8); released as model_derived |
| V054 | 3 | 5 | Calculated curves kept: curves "i" and "iii" are simulated, which only the caption says |
| V057 | 1 | 4 | Calculated curves kept: simulated, refined and difference curves, no legend in the panel |
| V008 | 1 | 2 | Inset curve kept: a background-corrected copy of the measurement inside an inset |
| V059 | 2 | 5 | Inset curves kept: three curves of a magnified inset |
| V055 | 1 | 3 | Plot frame placed on the inset: the crop had lost the panel's own x axis |
| V011 | 7 | 6 | Measured curve lost: its trace spanned less than 5° (Section S2.7) |

**Curve pairing.** For scoring, the operator's and the pipeline's curves of a panel are paired one to one. Pairs are made directly where each side has one curve (29 pairs), otherwise by label (74) or by the smallest total profile RMSE (22).

**Scoring.** Both curves of a pair are interpolated onto a 0.005° grid over their shared range and smoothed by a 0.25° quadratic Savitzky–Golay filter. After rescaling to 0–100, their peaks need a prominence of at least 5, and each is refined by a parabola. The Hungarian algorithm [18] pairs the two peak lists one to one so that the total distance is smallest, and pairs within 0.5° count as matched. A curve's peak-position difference is the median absolute difference of its matched peaks. Its profile RMSE uses the coarser curve's median point spacing as the grid step.

**Random pairing.** The random pairing swaps each pipeline curve for a random curve from another panel in one seeded draw. Of its 125 pairs, 111 overlap enough to score.

## S5.3. Synthetic Ground Truth

**Synthetic curves.** Each synthetic curve is sampled on a 0.02° grid and has four to eight pseudo-Voigt peaks at known positions. A pseudo-Voigt peak mixes a Gaussian and a Lorentzian peak of the same width. The first peak lies within the lowest 15% of the axis span, and the others sit at irregular multiples of its position. Peak heights fall and widths grow toward higher angle. Each curve also carries a broad stacking hump, a flat background and Gaussian noise.

**Rendering.** Ten rendering conditions of six panels each gave 60 panels and 222 curves. They were drawn with matplotlib [19] at 4.2 × 3.2 in, and stacked curves each span 0 to 100. The reference condition, Four curves, uses an offset of 115 between curves, 1.2-pt lines, 200 dpi, 2–30° with 5° ticks and noise 0.6. The other nine conditions each change one setting. They are One, Two and Six curves, Grayscale, Overlapping (offset 55), Thick lines (2.6 pt), Low resolution (110 dpi), Wide 2θ range (5–80°, 10° ticks) and Noisy (noise 2.4).

**Extraction.** The panels were extracted with the released code (Section S2.8), with true values in place of the LLM's outputs. Without OCR, every axis was a lone fit, and traces were not clipped to the calibrated window. The exact anchor has 400 points, and the coarse anchor has 60 points that scatter by 1.2% of the plot height.

**Scoring.** Scoring followed Section S5.2 on a 0.02° grid against the 1,206 true peaks. The signed error is the extracted minus the true peak position. With the exact anchor, 203 of 222 curves (91.4%) have a negative signed error (Figure S6).

**Origin of the shift.** Probe panels with two reference lines at known 2θ give the true position of each tick value. Tick detection lands exactly on the drawn marks at all four probe settings (0.00 px). Matplotlib draws the marks a mean of 0.08 to 0.61 px from their true positions (up to 1.02 px for single ticks). At the three settings that match a synthetic condition (Figure S6c), the resulting axis offset matches the synthetic signed error within 0.02°. This offset thus explains most of the −0.032° shift.

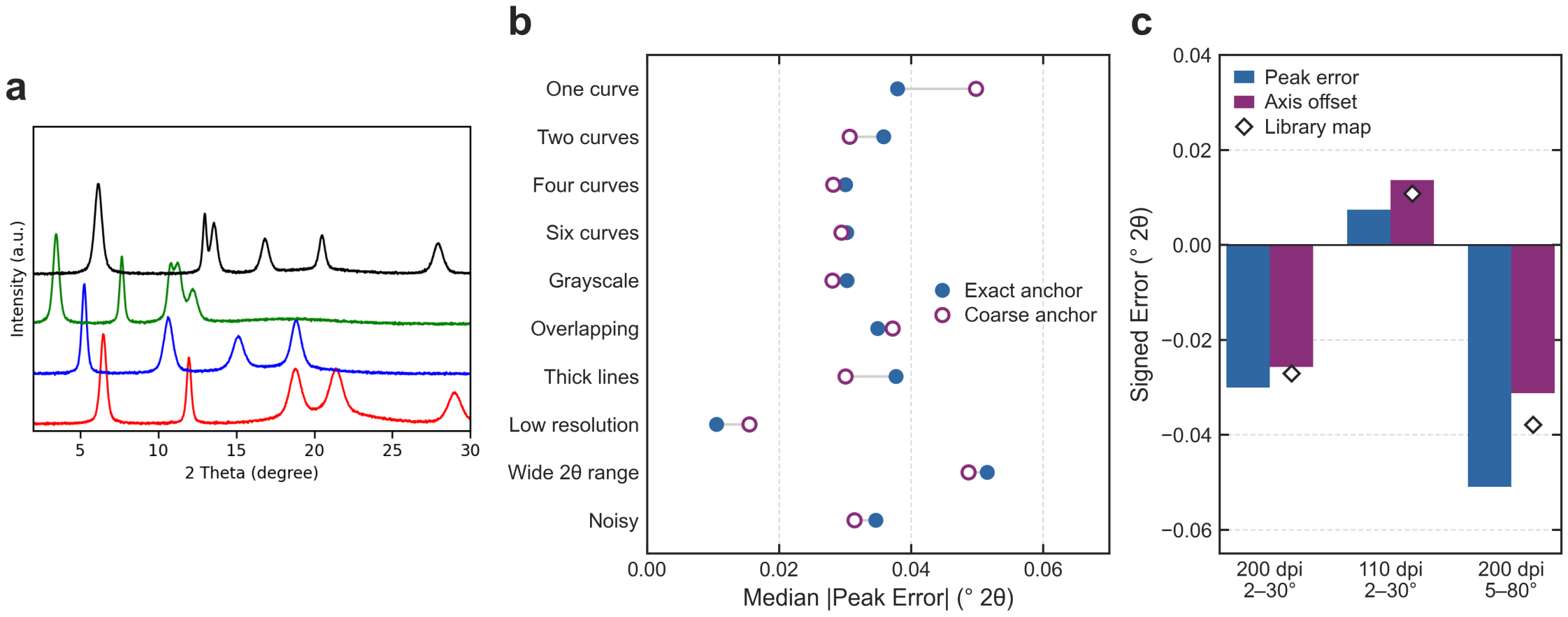

**Figure S6. Peak-position errors on the 60 synthetic panels of known truth.** (a) An Overlapping panel; (b) median absolute peak error per condition, exact (blue) and coarse (purple) anchor; (c) median signed error (blue) and axis offset from tick drawing (purple; diamond, matplotlib's map), which explains most of it.

**Both estimates.** The operator and the pipeline both calibrate on the drawn ticks. A tick-drawing offset therefore cancels in the hand comparison but enters the synthetic error, which is measured against the true curve. Both estimates leave out rejected curves, wrong links and curves that the protocol excludes but the pipeline kept (Table S17).

# Section S6. Applications of the Digitized Curves

This section describes two analyses of the digitized curves, the synthesis-parameter ladders (Section S6.1) and the recipe tests with their power analysis (Section S6.2).

## S6.1. Synthesis-Parameter Ladders

**Selection.** Ladders were found from the curve labels in six steps (Table S18). A label value is a number with a unit of time or temperature that does not directly follow a letter or hyphen. Captions are classified by keywords, and pseudo-three-dimensional plots are removed because their skewed axes distort 2θ. The 64 ladders hold 284 curves, each with at least 30 points and a value from its own label. Two of these ladders (8 curves) come from the 124 figures of Section S3.5.

**Table S18.** Selection of the synthesis-parameter ladders, in which each row applies one step to the figures left by the rows above it.

| Step | Figures | Papers |
|---|---|---|
| At least 3 values of one unit in the labels | 171 | – |
| Same label apart from the value, one material | 138 | 105 |
| Caption describes synthesis conditions | 69 | 55 |
| Material is a COF | 67 | 53 |
| Not a pseudo-three-dimensional plot | 66 | 52 |
| Parameter is reaction time or temperature | 64 | 51 |

**Measures.** Each curve is shifted down by its 5th percentile, scaled to a maximum of 1 and clipped at zero. Peaks are found with SciPy and need a prominence of at least 0.04 and a separation of at least 0.08°. The strongest low-angle peak is the most prominent peak at or below 15°, or anywhere if none lies there. $P$ is its prominence and $W$ its full width at half prominence, in degrees.

**Trends.** Each ladder's trend is the Spearman ρ [20] between parameter value and measure, with the sign reversed for width. This rank correlation describes how consistently the measure rises or falls with the parameter value. Each paper contributes the median ρ of its ladders, with zero counting as a tie.

**Tests.** A paper enters the test of a measure only if one of its ladders has that measure on at least three curves with different parameter values. This rule leaves 45 papers for $P$ and 51 for $W$. Papers are tested with a two-sided sign test [21], which checks whether positive and negative trends are equally common. Papers are also tested with a permutation test, which shuffles the values within each ladder 2,000 times and compares the observed split of papers with the shuffled ones.

**Results.** For $P$, the trend was positive in 31 papers, negative in 13 and tied in 1 (sign test p = 0.0096, permutation p = 0.0045). For $W$, the counts were 26, 24 and 1 (p = 0.89 and 0.83). The median per-paper ρ was +0.45 for prominence and +0.10 for width (Figure S7a).

**Robustness.** At the reported settings of the robustness check (Table S19; Figure S7b), removing any single paper leaves the prominence p between 0.0054 and 0.014. Five positive papers must be removed before p exceeds 0.05.

**Table S19.** Robustness check of the ladder sign test over 2,880 combinations of seven settings fixed by the written plan (Section S1.4), with the reported value of each in bold. The lower block gives results as prominence / width.

| Setting varied | Values tried |
|---|---|
| Minimum peak prominence (0–1 scale) | 0.02, 0.03, **0.04**, 0.06, 0.08 |
| Upper end of the low-angle region (° 2θ) | 10, 12.5, **15**, 20 |
| Percentile subtracted as baseline | 0, 2, **5**, 10 |
| Minimum separation between peaks (° 2θ) | 0.04, **0.08**, 0.15 |
| Minimum number of parameter values per ladder | **3**, 4 |
| How a paper's ladders are combined | **median**, mean, majority vote of signs |
| Papers whose combined value is zero | **dropped**, counted against the trend |
| **Result over the 2,880 combinations** | **Prominence / width** |
| $p < 0.05$ in the declared direction | 95.5% / 0% |
| Median p (range) | 0.0039 (0.00007 to 0.10) / 0.78 (0.19 to 1.0) |

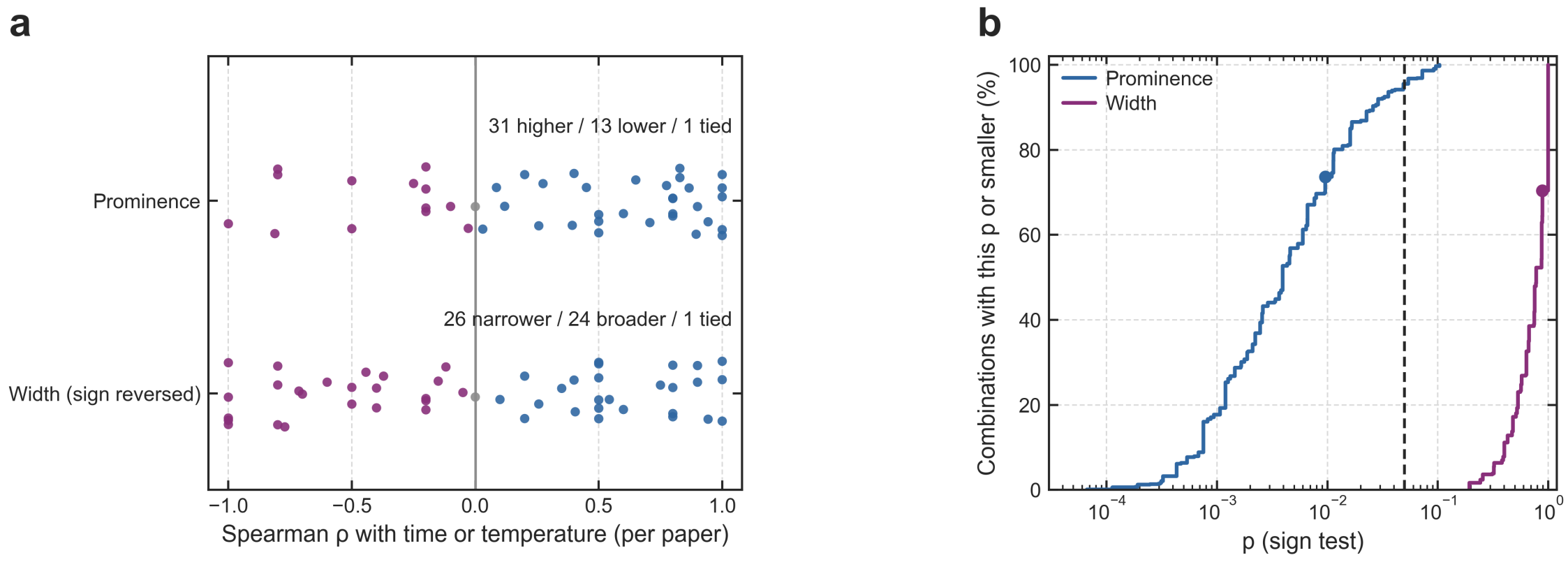

**Figure S7. Trends of peak prominence and width across the synthesis-parameter ladders.** (a) Per-paper median Spearman ρ, prominence and sign-reversed width; (b) share of the 2,880 setting combinations with sign-test $p \leq x$ (dashed, $p = 0.05$; dots, reported settings).

## S6.2. Recipe Analyses and Statistical Power

**Linking.** The recipe tests use the 1,953 of 11,445 accepted curves (17.1%) that carry a high-confidence link to a record (Table S4). These links reach 970 records that name a framework class, from 543 papers.

**Tests.** Six recipe variables and three ranked outcomes ($W$, $P$ and the number of peaks) give 18 tests with one value per paper (Table S20). The statistic is $\Delta R^2$, the gain in $R^2$ over a linear model of framework class. Its p value comes from 2,000 within-class shuffles of the category. By the plan's decision rule, a test is positive if $\Delta R^2$ is at least 0.02 and p is at most 0.05/24, the Bonferroni level of the written plan. The plan's six other tests used an outcome that does not come from the curves and are not reported here.

**Results.** Only activation method on the number of peaks passed this rule. It failed when only chemically named categories were kept ($\Delta R^2$ 0.0162, $p = 0.023$, 444 papers).

**Table S20.** $\Delta R^2$ beyond framework class for each of the 18 recipe tests, with the permutation p value in parentheses. The Papers column gives the range over the three outcomes, and * marks $p \leq 0.0021$, the Bonferroni level.

| Recipe variable | Papers | Width | Prominence | Number of peaks |
|---|---|---|---|---|
| Solvent | 499–502 | 0.0390 (0.0035) | 0.0295 (0.036) | 0.0280 (0.042) |
| Modulator | 469–472 | 0.0220 (0.0095) | 0.0194 (0.031) | 0.0128 (0.12) |
| Temperature | 430–432 | 0.0183 (0.10) | 0.0076 (0.50) | 0.0158 (0.14) |
| Reaction time | 498–501 | 0.0191 (0.015) | 0.0073 (0.31) | 0.0136 (0.069) |
| Synthesis method | 518–521 | 0.0236 (0.016) | 0.0148 (0.18) | 0.0099 (0.39) |
| Activation method | 484–487 | 0.0089 (0.32) | 0.0087 (0.37) | 0.0447 (0.0005)* |

**Power.** A test's power is its chance of detecting a real effect of a given size. Effects of known size are planted in three patterns into the real papers and categories of each test. The minimum detectable effect of a test is the smallest $\Delta R^2$ that it detects 80% of the time at the Bonferroni level, under its least favorable pattern.

**Detectable effects.** The median minimum detectable effect over the 18 tests is 0.050 (range 0.040 to 0.062). Every observed $\Delta R^2$, at most 0.0447, lies below its own test's minimum. Detecting an effect of 0.02 with 80% power would need at least 1,015 papers per test (median), 2.1 times those available.

# Section S7. The COF-PEAK Database

## S7.1. Files and Access

**Files.** The release holds all 11,445 accepted curves in three gzip-compressed CSV files. The file curves.csv.gz has one row per curve, points.csv.gz one row per point and papers.csv.gz one row per paper. The files are attached under CC BY 4.0 to the releases of the GitHub repository https://github.com/AaronG211/pxrd_PEAK. A README describes all their columns and gives cautions for use.

**First peak.** A curve's first peak is its lowest-angle peak that passes five checks (signal to noise, persistence, prominence, width and distance from the range start; thresholds in the code, Section S1.1). Its status is ok when it rises at least 5 times the noise level above the local baseline, with the noise level estimated from the median absolute deviation. The status is low_confidence when the peak rises 3.5 to 5 times the noise level, or when a lower candidate was cut off at the start of the range. A curve without such a peak gets no first-peak value, and its status is no_bragg_peak or truncated_at_window_start.

**Website and API.** The COF-PEAK website (https://pxrd-peak.vercel.app) and its read-only API serve the same records under the same IDs. The API needs no key and names some fields differently (documentation at https://pxrd-peak.vercel.app/docs/api).

**Source articles.** Like the website, the files hold no figure image from the source articles. A curve's source figure is reached through the article's DOI. The website and the API reproduce no text from the source articles beyond bibliographic data, and they link to each article's DOI instead.

## S7.2. Admission Classes

**Offline checks.** After the run, each accepted curve was retested and given one of five admission classes. The retest applied eight offline trace checks, whose thresholds are in the code (Section S1.1). Four checks read the shape of the stored curve, and four read the ink-snapping evidence, such as precision and snap rate. A single failed curve keeps its whole figure out of clean. The reason is that a figure's curves share one 2θ axis, and a failed trace can signal a misread panel.

**Class order.** Each curve takes the first class whose condition it meets (Table S21). The curves of each class are counted in Table S22.

**Table S21.** Conditions of the admission classes, in the order they are tested.

| Class | Condition |
|---|---|
| model_derived | The keyword filter of Section S2.2, rerun on the stored labels, marks the curve as model-derived |
| shape_flagged | The curve failed an offline trace check |
| clean | No curve of its figure failed a check, and the figure is not one of the 124 of Section S3.5 |
| axis_verified | The curve passed every check, another curve of its figure failed one, and the axis was calibrated by consensus |
| axis_unconfirmed | As for axis_verified, on a lone-fit or arbitrated axis; also every curve on the 124 figures that passes every check |

**Table S22.** Curves in each admission class by the calibration route of their figure's axis. Figures and papers count those that hold at least one curve of the class, so their column sums exceed the totals.

| Class | Consensus | Lone-fit rule | Axis arbitration | Curves | Figures | Papers |
|---|---|---|---|---|---|---|
| clean | 6,410 | 1,099 | 49 | 7,558 | 3,597 | 1,964 |
| axis_verified | 1,672 | 0 | 0 | 1,672 | 581 | 468 |
| shape_flagged | 961 | 319 | 59 | 1,339 | 949 | 732 |
| axis_unconfirmed | 0 | 435 | 237 | 672 | 276 | 220 |
| model_derived | 160 | 39 | 5 | 204 | 108 | 76 |
| All | 9,203 | 1,892 | 350 | 11,445 | 4,664 | 2,370 |

**Axis caveats.** The clean class does not test the axis, since 1,099 of its curves lie on lone-fit axes and 49 on arbitrated axes. None of the 241 accepted curves on the 124 figures of Section S3.5 is clean. The 206 that pass every check are axis_unconfirmed, 32 are shape_flagged and 3 are model_derived.

**Recommended use.** The recommended default set is the clean and axis_verified curves (9,230 curves, 4,178 figures, 2,203 papers). Comparisons of absolute 2θ across papers should drop arbitrated axes (9,181 curves, 4,150 figures, 2,191 papers). The strictest choice for such comparisons keeps consensus axes only (8,082 curves, 3,631 figures, 1,972 papers). The

shape_flagged and axis_unconfirmed curves should be used only after checking the source figure through the article's DOI. The model_derived curves should be used only as calculated patterns.

## S7.3. Random Sample of Released Curves

Figure S8 shows curves drawn at random from the recommended default set. It gives a view of typical records, with no curve chosen by hand.

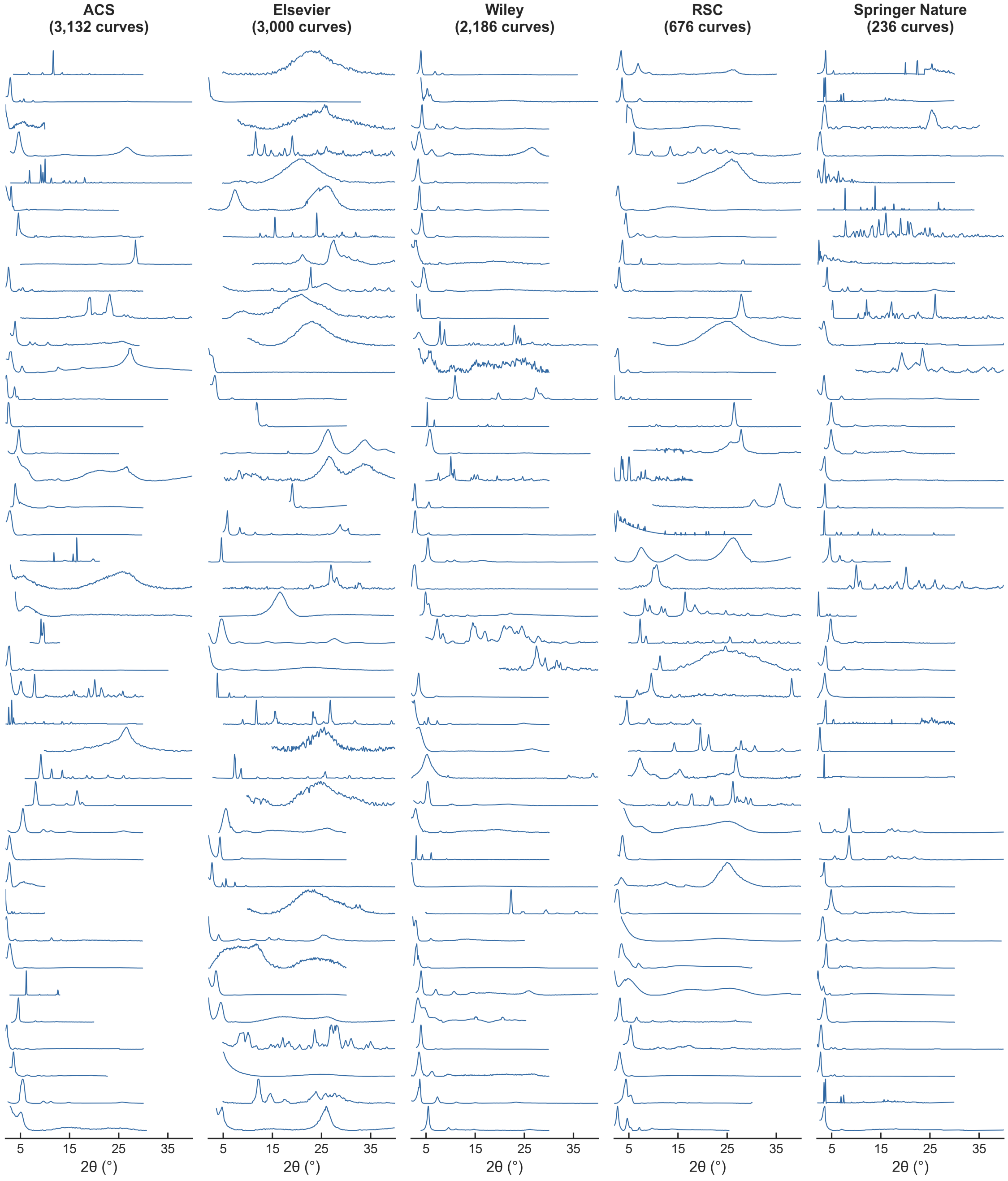

**Figure S8. Random sample of released curves.** For each publisher, 40 curves drawn at random (seed 20261029) from the default set (Section S7.2), replotted from the stored points with an offset and scaled to 0–1 between 2° and 40°. One Springer Nature curve lies wholly above 40°.

# References

[1] Y. Du *et al.*, "PP-OCR: A Practical Ultra Lightweight OCR System." arXiv, 2020. doi: 10.48550/ARXIV.2009.09941.

[2] P. Virtanen *et al.*, "SciPy 1.0: Fundamental Algorithms for Scientific Computing in Python," *Nat. Methods*, vol. 17, no. 3, pp. 261–272, Feb. 2020, doi: 10.1038/s41592-019-0686-2.

[3] C. Birkle, D. A. Pendlebury, J. Schnell, and J. Adams, "Web of Science as a Data Source for Research on Scientific and Scholarly Activity," *Quant. Sci. Stud.*, vol. 1, no. 1, pp. 363–376, Feb. 2020, doi: 10.1162/qss_a_00018.

[4] L. Foppiano, G. Lambard, T. Amagasa, and M. Ishii, "Mining Experimental Data from Materials Science Literature with Large Language Models: An Evaluation Study," *Sci. Technol. Adv. Mater. Methods*, vol. 4, no. 1, p. 2356506, Jul. 2024, doi: 10.1080/27660400.2024.2356506.

[5] B. Efron, "Bootstrap Methods: Another Look at the Jackknife," *Ann. Stat.*, vol. 7, no. 1, pp. 1–26, Jan. 1979, doi: 10.1214/aos/1176344552.

[6] C. A. Field and A. H. Welsh, "Bootstrapping Clustered Data," *J. R. Stat. Soc. Ser. B Stat. Methodol.*, vol. 69, no. 3, pp. 369–390, May 2007, doi: 10.1111/j.1467-9868.2007.00593.x.

[7] P. H. C. Eilers and H. F. M. Boelens, "Baseline Correction with Asymmetric Least Squares Smoothing," Leiden University Medical Centre, Leiden, The Netherlands, 2005.

[8] A. Savitzky and M. J. E. Golay, "Smoothing and Differentiation of Data by Simplified Least Squares Procedures," *Anal. Chem.*, vol. 36, no. 8, pp. 1627–1639, Jul. 1964, doi: 10.1021/ac60214a047.

[9] J. Maschita, T. Banerjee, and B. V. Lotsch, "Direct and Linker-Exchange Alcohol-Assisted Hydrothermal Synthesis of Imide-Linked Covalent Organic Frameworks," *Chem. Mater.*, vol. 34, no. 5, pp. 2249–2258, Feb. 2022, doi: 10.1021/acs.chemmater.1c04051.

[10] Y. Lan *et al.*, "Materials Genomics Methods for High-Throughput Construction of COFs and Targeted Synthesis," *Nat. Commun.*, vol. 9, no. 1, p. 5274, Dec. 2018, doi: 10.1038/s41467-018-07720-x.

[11] M. Everingham, L. Van Gool, C. K. I. Williams, J. Winn, and A. Zisserman, "The Pascal Visual Object Classes (VOC) Challenge," *Int. J. Comput. Vis.*, vol. 88, no. 2, pp. 303–338, 2010, doi: 10.1007/s11263-009-0275-4.

[12] A. G. Seco de Herrera, R. Schaer, S. Bromuri, and H. Müller, "Overview of the ImageCLEF 2016 Medical Task," in *CLEF 2016 Working Notes*, in CEUR Workshop Proceedings, vol. 1609. 2016, pp. 219–232. [Online]. Available: https://ceur-ws.org/Vol-1609/16090219.pdf

[13] S. Tsutsui and D. J. Crandall, "A Data Driven Approach for Compound Figure Separation Using Convolutional Neural Networks," in *2017 14th IAPR International Conference on Document Analysis and Recognition (ICDAR)*, IEEE, Nov. 2017, pp. 533–540. doi: 10.1109/icdar.2017.93.

[14] W. Jiang, E. Schwenker, T. Spreadbury, N. Ferrier, M. K. Y. Chan, and O. Cossairt, “A Two-Stage Framework for Compound Figure Separation,” in *2021 IEEE International Conference on Image Processing (ICIP)*, IEEE, Sep. 2021, pp. 1204–1208. doi: 10.1109/icip42928.2021.9506171.

[15] S. Lloyd, “Least Squares Quantization in PCM,” *IEEE Trans. Inf. Theory*, vol. 28, no. 2, pp. 129–137, Mar. 1982, doi: 10.1109/tit.1982.1056489.

[16] B. Lukose, A. Kuc, and T. Heine, “The Structure of Layered Covalent-Organic Frameworks,” *Chem. Eur. J.*, vol. 17, no. 8, pp. 2388–2392, Jan. 2011, doi: 10.1002/chem.201001290.

[17] A. Rohatgi, *WebPlotDigitizer*. Automeris. [Online]. Available: https://automeris.io

[18] H. W. Kuhn, “The Hungarian Method for the Assignment Problem,” *Nav. Res. Logist. Q.*, vol. 2, no. 1–2, pp. 83–97, Mar. 1955, doi: 10.1002/nav.3800020109.

[19] J. D. Hunter, “Matplotlib: A 2D Graphics Environment,” *Comput. Sci. Eng.*, vol. 9, no. 3, pp. 90–95, 2007, doi: 10.1109/mcse.2007.55.

[20] C. Spearman, “The Proof and Measurement of Association Between Two Things,” *Am. J. Psychol.*, vol. 15, no. 1, pp. 72–101, Jan. 1904, doi: 10.2307/1412159.

[21] W. J. Dixon and A. M. Mood, “The Statistical Sign Test,” *J. Am. Stat. Assoc.*, vol. 41, no. 236, pp. 557–566, Dec. 1946, doi: 10.1080/01621459.1946.10501898.